\documentclass[wcp]{jmlr}
\usepackage{natbib}

\usepackage{longtable}

\usepackage{booktabs}
\usepackage{amsmath,amssymb, mathrsfs,empheq}
\usepackage{algorithm}
\usepackage{algorithmic}
\usepackage{bm}
\usepackage{multirow}

\title[Learning GB-RBM with DC]{Learning Gaussian-Bernoulli RBMs using Difference of Convex Functions Optimization}

  \author{\Name{Vidyadhar Upadhya} \Email{vidyadhar@ee.iisc.ernet.in}\\
  \addr Electrical Engineering Dept, Indian Institute of Science, Bangalore 
  \AND
  \Name{P. S. Sastry} \Email{sastry@ee.iisc.ernet.in}\\
  \addr Electrical Engineering. Dept, Indian Institute of Science, Bangalore
 }

\newcommand{\beq}{\begin{equation}}
\newcommand{\eeq}{\end{equation}}
\newcommand{\beqa}{\begin{eqnarray}}
\newcommand{\eeqa}{\end{eqnarray}}
\newcommand{\ben}{\begin{enumerate}}
\newcommand{\een}{\end{enumerate}}

\newcommand{\diag}{\mathrm{diag}}

\newcommand{\llk}{\mathcal{L}}
\newcommand{\Real}{\mathbb{R}}

\newcommand{\rb}{\right)}
\newcommand{\lb}{\left(}
\newcommand{\rsq}{\right]}
\newcommand{\ls}{\left[}

\newcommand{\vecx}{\mathbf{x}}

\newcommand{\vecv}{\mathbf{v}}
\newcommand{\vecu}{\mathbf{u}}
\newcommand{\veca}{\mathbf{a}}
\newcommand{\vecs}{\mathbf{s}}
\newcommand{\vecb}{\mathbf{b}}
\newcommand{\vecc}{\mathbf{c}}
\newcommand{\vecw}{\mathbf{w}}
\newcommand{\vech}{\mathbf{h}}

\newcommand{\vecbeta}{\mathbf{\beta}}
\newcommand{\vecone}{\mathbf{1}}

\newcommand{\Ex}{\mathbb{E}}

\newcommand{\sth}{\tilde{\theta}}
\newcommand{\sv}{\tilde{\vecv}}

\newcommand{\non}{\nonumber}

\def  \wrt{\textit{w.r.t. }}
\date{}

\begin{document}
\maketitle


\begin{abstract}
The Gaussian-Bernoulli restricted Boltzmann machine (GB-RBM) is a useful generative model that captures 
meaningful features from the given $n$-dimensional continuous data.
The difficulties associated with learning GB-RBM are reported extensively in earlier studies.
They indicate that the training of the GB-RBM using the current standard algorithms, namely,  contrastive divergence (CD) and persistent contrastive divergence (PCD), needs a
carefully chosen small learning rate to avoid divergence which, in turn, results in slow learning. 
In this work, we alleviate such difficulties by showing that the negative log-likelihood for a GB-RBM can be expressed as a difference of convex functions 
if we keep the variance of the conditional distribution of visible units (given hidden unit states) and the biases of the visible units, constant. Using this, we propose a  
stochastic {\em difference of convex functions} (DC) programming (S-DCP) algorithm for learning the GB-RBM.  We present extensive empirical studies on several benchmark datasets to 
validate the performance of this S-DCP algorithm. 
It is seen that S-DCP  is better than the CD and PCD
algorithms in terms of speed of learning and the quality of
the generative model learnt.
\end{abstract}



%

\section{Introduction}

Probabilistic generative models are good for extracting meaningful representations from the data by learning high-dimensional multi-modal distributions
in an unsupervised manner. 
The restricted Boltzmann machine (RBM) is one such generative model which 
was originally proposed for modeling distributions of 
binary data variables (called visible units) with the 
help of binary latent variables (called hidden units) \citep{smolensky1986information,freund1994unsupervised,hinton2002training}.  The RBM is found to be effective for unsupervised representation 
learning in several applications and has been used as a building block for models such as deep belief networks, deep Boltzmann machines~\citep{Hinton06,salakhutdinov2009deep,6763041,8941264,8038860} and 
also in combination with the Convolutional Neural Networks \citep{ZHANG2018156}. It can also be used as a 
discriminative model \citep{Hinton06,ZHANG2019104911}. 

The RBM (like the Boltzmann machine) uses a suitable energy function to represent the probability distribution as a Gibbs distribution. 
The distinguishing feature of an RBM (compared to the Boltzmann machine) is that
the hidden units are conditionally independent, conditioned on all the visible units and vice versa. These conditional distributions are enough to define an energy function and hence, the joint 
distribution. The normalizing constant for the distribution, known as the partition function, is computationally intractable. However, one can efficiently 
sample from the distribution 
through Markov Chain Mote Carlo (MCMC) methods. 

In the original RBM, both visible and hidden units are assumed to be binary. Hence, the model is also called Bernoulli-Bernoulli RBM (BB-RBM). It is to be noted that the visible units, being observable, 
constitute the data variables. 
To model continuous valued data, the binary visible units must be replaced with  continuous ones though the hidden units can still 
remain binary. 
One possibility is to model the distribution of visible units conditioned on the hidden units as a Gaussian.
The resulting model is called Gaussian-Bernoulli RBM (GB-RBM) \citep{NIPS2004_2672,Hinton504}. 
In this work, we address the important problem of finding efficient algorithms for learning GB-RBMs.

The usual approach to learning an RBM is by minimizing the KL divergence between the data and the model distributions
which is equivalent to finding the maximum likelihood estimate of the parameters of the model.
The maximum likelihood estimates have no closed form solution and the parameters are learnt using iterative algorithms to minimize the negative log likelihood. 
The log-likelihood gradient requires calculating an expectation 
\textit{w.r.t.} the model distribution which is computationally expensive due to the presence of the intractable partition function. 
Therefore, this expectation is estimated using samples obtained from the model distribution using MCMC methods.
This strategy is followed in the popular Contrastive Divergence (CD) algorithm and its variant, the Persistent Contrastive 
Divergence (PCD); the only algorithms currently used for 
learning GB-RBMs. 
However, the estimated gradient may have a large variance due to limitations in the MCMC procedure and
can cause  these stochastic gradient descent based algorithms to diverge sometimes~\citep{cho}. The issue of divergence is more 
prominent in the case of GB-RBM compared to BB-RBM~\citep{DBLP:journals/corr/WangMW14}.

It was shown in \cite{NIPS2007_3313} that learning GB-RBM with a sparse 
penalty provides meaningful representations for natural images. Although it was empirically observed that simple mixture 
models can outperform the GB-RBM models in terms of the 
likelihood estimation \citep{Theis:2011:LDB:1953048.2078204}, with the re-parameterization of 
energy function and improved learning algorithm in \cite{cho} the 
GB-RBM model was found to extract more meaningful representations. 
However, learning GB-RBM is a challenging task and 
difficulties associated with learning are reported extensively 
in the literature \citep{citeulike:7491128, Theis:2011:LDB:1953048.2078204, cho, DBLP:journals/corr/WangMW14,10.1371/journal.pone.0171015}. 
For instance, the analysis in \cite{DBLP:journals/corr/WangMW14} showed that the failures in 
learning GB-RBMs are due to the inefficiency of training algorithms rather than any inherent limitations of the model itself. 
Several training recipes 
which make use of some knowledge of the data distribution were 
also suggested. 
They also showed that the features learnt on benchmark 
datasets are comparable to those learnt using sparse coding 
and Independent Component Analysis (ICA). This claim is 
theoretically justified in \cite{KARAKIDA201678} where they showed that the GB-RBM, learnt using an exact maximum likelihood or an approximate maximum likelihood with CD estimates, extracts independent components at one of 
their stable fixed points. In addition, the difficulties to capture 
the geometric structure of the data manifold by GB-RBMs 
was overcome using approaches where the local manifold 
structure of the data is captured through graph regularization 
(GraphRBM) \citep{7927417} and using the guidance from the available 
label information (pcGRBM) \citep{8444749}. 

Currently, the CD and PCD algorithms are the standard algorithms for learning GB-RBM models.
They require a carefully chosen 
learning rate (which is usually small) so as to avoid divergence 
of the log-likelihood. The small value of the learning rate results in slow convergence. 
In addition, one also needs careful parameter initialization and 
restriction of the gradient during the update, for avoiding the 
divergence of the log-likelihood. Thus, there is a need for an 
efficient algorithm for learning GB-RBMs.

In this work, we show that the log-likelihood function of a GB-RBM can be expressed as a difference of two convex functions under the assumption that the conditional distribution of visible units is 
Gaussian with unit variance. There are efficient algorithms for optimizing difference of convex functions and this strategy has been used earlier for 
learning BB-RBMs \citep{carlson2015stochastic,pmlr-v54-nitanda17a,pmlr-v77-upadhya17a}. 
We adopt the stochastic difference of convex functions programming (S-DCP) \citep{pmlr-v77-upadhya17a} algorithm to learn GB-RBMs.
Through extensive empirical studies using benchmark datasets, we show that the S-DCP algorithm is more efficient compared to the existing CD and PCD algorithms for learning GB-RBMs. The main contributions of this
work are as follows.

We show that if the conditional distribution of the visible units is Gaussian with fixed variance, then the log likelihood function for GB-RBMs can be expressed as a difference of two 
convex functions. We then show how we can employ a two loop stochastic gradient descent algorithm (called S-DCP) to learn the GB-RBM and demonstrate through empirical studies  that this algorithm is more efficient than CD and PCD. Though we fix the variance of the conditional distribution of the visible units in our algorithm, we still efficiently learn good representations. 
Further, we modify the S-DCP algorithm to accommodate variance update and show that the 
resulting method is also more efficient than CD  and PCD.
We show that the proposed algorithm is better in terms of the speed of learning, achieved log-likelihood, ability to extract meaningful features and to generate high quality samples. 


The rest of the paper is organized as follows. In section \ref{sec:background}, 
we briefly describe the RBM model and the maximum likelihood approach to learn a GB-RBM. 
We show that the GB-RBM log-likelihood is a difference of convex functions and  describe the 
DC programming approach in section \ref{sec:ll_convex}.
In section \ref{sec:experiments}, we describe the 
experimental setup and present the  
results. We conclude the paper in 
section~\ref{sec:conclusions}.

\section{Background}\label{sec:background}
The RBM is a two layer recurrent neural network, in which $m$ visible stochastic units (denoted as $\vecv$) in one layer are connected to $n$ hidden stochastic units (denoted as $\vech$) in the other 
 \citep{6302931_2,freund1994unsupervised,hinton2002training}. There are no connections among the units within each layer and the connections between the two layers are undirected.
The weight of the connection between the $i^{\text{th}}$  visible unit and the $j^{\text{th}}$ hidden unit is denoted as $w_{ij}$.
The bias for the $i^{\text{th}}$ visible unit and the $j^{\text{th}}$ hidden unit are denoted as $b_i$ and $c_j$, respectively.
Let $\theta=\{\vecw\in\Real^{n\times m},\vecb\in\Real^{m},\vecc\in\Real^{n}\}$ denote the parameters of the model. 
RBM represents probability distribution over the states of its units as, 
\beq
 p(\vecv,\vech\vert\theta)=e^{-E(\vecv,\vech;\theta)}/Z(\theta),\label{eq:llk_eq}
 \eeq
 where $Z(\theta)$ 
is the normalizing constant, called the 
 partition function and 
$E(\vecv,\vech;\theta)$ is the energy function. 
The bipartite connectivity structure in RBM implies conditional independence of visible units conditioned on all hidden units and vice-versa, i.e.,  
\beq
p(\vecv\vert\vech,\theta)=\prod_{i=1}^m p(v_i\vert \vech,\theta),\quad\,\, p(\vech\vert\vecv,\theta)=\prod_{i=1}^n p(h_i\vert \vecv,\theta).\non
\eeq

\subsection{Bernoulli-Bernoulli Restricted Boltzmann Machines}
The RBM was originally proposed for binary visible and hidden units ($\vecv \in \{0,1\}^m$ and $\vech \in \{0,1\}^n$). In the BB-RBM the conditional distributions are prescribed as,
\beqa
p(v_i=1\vert\vech,\theta)&=& \text{sigmoid}\lb b_i +\sum_{j=1}^n w_{ij} h_j\rb\non\\
p(h_j=1\vert\vecv,\theta) &=&\text{sigmoid}\lb c_j+\sum_{i=1}^m w_{ij} v_i\rb\non
\eeqa
where, $\text{sigmoid}(x) = 1/(1+e^{-x})$ is the logistic sigmoid function.
Using these, we obtain the following energy function.
\beqa
E(\vecv,\vech;\theta)=-\sum_{i,j}w_{ij} v_i \, h_j\, -\sum_{i=1}^{m} b_i\, v_i-\sum_{j=1}^{n} c_j\, h_j.
\eeqa
The data distribution represented by an RBM is that of the visible units which is obtained by marginalizing from  eq. \eqref{eq:llk_eq}.
One can show that any distribution over $\{ 0, \; 1\}^m$ can be approximated by a BB-RBM with ($m$ visible units and) sufficient number of hidden units~\citep{MONTUFAR2017531,8475002}. 

\subsection{Gaussian-Bernoulli Restricted Boltzmann Machines}
The BB-RBM can represent a data distribution only if the data variables are binary. 
The natural extension of the ~{BB-RBM} is to choose continuous visible units and binary hidden units, i.e., $\vecv\in\mathcal{R}^{m}$ and $\vech\in\{0,1\}^{n}$.
In this case, the conditional distribution of the units are prescribed as, 
\beqa
p(v_i\vert\vech,\theta)&=& \mathcal{N}\lb v_i\vert\,\, b_i+\sum\limits_j w_{ij} \, h_j, \; \sigma_i^2\rb\non\\
p(h_j=1\vert \vecv,\theta)&=& \text{sigmoid}\lb c_j+\sum\limits_i w_{ij} \, \frac{v_i}{\sigma_i^2}\rb\label{eq:rbm_cond}
\eeqa
where, $\mathcal{N}(.\vert \mu,\sigma^2)$ denotes Gaussian probability density function with mean $\mu$ and
variance $\sigma^2$ \cite{cho}. This RBM model is called the Gaussian-Bernoulli RBM (GB-RBM). 

The choice of the following energy function results in the desired conditional distributions as in eq. \eqref{eq:rbm_cond} ,
\beq
E(\vecv,\vech\vert\theta)=\sum_i \frac{(v_i-b_i)^2}{2\sigma_i^2}-\sum_{i,j} \frac{w_{ij} v_i \, h_j}{\sigma_i^2} -\sum_j c_j\,h_j\non
\eeq
where, $\sigma_i^2$ is the variance associated with the $i^{\text{th}}$ Gaussian visible unit \citep{cho}. 
Using this energy function with $\sigma_i^2=1,\forall i$,  we obtain (from eq. \eqref{eq:llk_eq}) the partition function $Z(\theta)$ as,

\beq
Z(\theta)=\sum_\vech\int_{\Real^m} e^{-\sum\limits_i \frac{(v_i-b_i)^2}{2}+\sum\limits_{i,j} w_{ij} v_i \, h_j +\sum\limits_j c_j\,h_j} \mathbf{d}\vecv.\non
\eeq
\subsection{Maximum Likelihood Learning}\label{sec_ML}
One can learn
the GB-RBM parameters to represent the data distribution, given some training data, 
 by maximizing the log-likelihood. For simplicity of notation, let us consider a single training sample ($\vecv$). Then log-likelihood is given by,
\beqa
\llk (\theta\vert \vecv)&=&\log\,p(\vecv\vert\theta)
 = \log\,\sum_\vech p(\vecv,\vech\vert\theta)\non\\
& \triangleq & ( g(\theta,\vecv) - f(\theta)) \label{eq:ll_base}
\eeqa
where,  
\beqa
g(\theta,\vecv)&=& \log\,\sum_\vech e^{-E(\vecv,\vech:\theta)}\non\\
f(\theta)&=& \log\,Z(\theta)=\log \,\sum_{\vech}\int_{\Real^m} e^{-E(\vecv,\vech;\theta)} \mathbf{d}\vecv.\label{eq:f_g_def}
\eeqa
The required RBM parameters can be found as 
\beq
\theta^*=\arg \max_\theta \llk (\theta\vert \vecv)= \arg \max_\theta \,\,( g(\theta,\vecv)- f(\theta)).\non
\eeq
Since there is no closed form solution, iterative gradient ascent procedure is used as,
\beq
\theta^{t+1}=\theta^{t}+ \eta\ls \nabla_\theta\,g(\theta,\vecv)- \nabla_\theta \,f(\theta) \rsq_{\theta=\theta^t}.\non
\eeq
It is easily shown that the gradient of $g$ and $f$ are given by, 
\beqa
\nabla_\theta\,g(\theta,\vecv)&=&
-\Ex_{p(\vech\vert\vecv;\theta)}\ls \nabla_\theta\,E(\vecv,\vech;\theta)\rsq\non\\
\nabla_\theta \,f(\theta)
&=&-\Ex_{p(\vecv,\vech;\theta)}\ls\nabla_\theta\, E(\vecv,\vech;\theta)\rsq.\label{eq:loglik_grad_f}
\eeqa
Here, $ \Ex_q$ denotes the expectation \wrt the distribution $q$. 
Since $\dfrac{\partial E(\vecv, \vech; \theta)}{\partial w_{ij}} = - v_ih_j$, we have,
\beqa
\frac{\partial g(\theta,\vecv)}{\partial w_{ij}} &= & \sum\limits_\vech v_i \, h_j \, p(\vech\vert\vecv,\theta) =  v_i \,\,p(h_j=1 \vert \vecv,\theta) \non \\
& = & v_i \,\,\sigma\lb c_j+\sum_{i=1}^m w_{ij} v_i\rb.\non
\eeqa
Similar to this case, the partial derivatives of 
$g(\theta,\vecv)$ \wrt $b_i$ and $ c_j$ yield expressions that are analytically easy to evaluate.

In contrast, it is computationally very expensive to calculate $\nabla_\theta \,f(\theta)$ due to the expectation in eq. \eqref{eq:loglik_grad_f}. 
For example, 
\beqa
\frac{\partial f(\theta)}{\partial w_{ij}} &=& \int_{\Real^m} \sum\limits_{\vech} v_i \, h_j \, p(\vecv,\vech\vert\theta)\,\, d\vecv \non\\
&=& \int_{\Real^m} v_i \,\,p(h_j=1 \vert \vecv,\theta) \,p(\vecv \vert \theta) \,d\vecv\non
\eeqa
which is computationally hard. 
Hence, MCMC sampling methods are used to estimate the expectation to obtain $\nabla_\theta\,f$. 

Due to the bipartite connectivity structure of the model, MCMC sampling can be efficiently implemented through
block Gibbs sampler. 
This is followed in
the contrastive divergence (CD) algorithm~\citep{hinton2002training} which is currently the most popular algorithm for learning GB-RBMs. In the CD($k$) algorithm, the Markov chain is run for $k$ block 
Gibbs sampling iterations and the $k^\text{th}$ state is used for a single-sample approximation of the expectation. 
The training samples are used to initialize the Markov chain for every mini-batch.
In a variant of this algorithm, called persistent contrastive divergence (PCD) \citep{tieleman2008training}, the Markov chain is initialized with 
the $k^\text{th}$ state of the previous iteration\textquoteright s Markov chain.

Currently, CD and PCD are the standard algorithms used to train GB-RBMs. However, these algorithms are 
 sensitive to  hyperparameters and require a carefully chosen learning rate. Generally, the learning rate needs to be very small to avoid divergence and hence, these algorithms learn very slowly. 
This issue is more prominent in GB-RBMs 
for reasons described in \cite{hinton2010practical,cho}. 
As explained earlier, our objective is to train the GB-RBMs
efficiently using
algorithms that optimize the difference
of convex functions.

\section{Difference of Convex Functions Formulation}\label{sec:ll_convex}
In this section, we show that the $f$ and $g$ functions given by eq. \eqref{eq:f_g_def}  are convex \wrt  $\vecw$ and $\vecc$ under the
assumption of fixed variance, $\sigma_i^2$. 
This implies that the log-likelihood function (considered as function of $\vecw,\vecc$), given by eq. \eqref{eq:ll_base}, is a difference of two convex
functions.
This property can be exploited to design efficient optimization algorithms through the \textit{Difference of Convex} (DC) programming approach. 
Here we choose $\sigma_i^2 =1, \forall i$, for the sake of convenience.
 It is easy to see that the result holds for any fixed value of $\sigma^2_i$.
\subsection{Convexity of $g(\theta,\vecv)$}\label{sub_sec_cvx_g}
The function $g(\theta,\vecv)$ is given by,
\beq
g(\theta,\vecv)= \log \sum_\vech e^{-\sum_i \frac{(v_i-b_i)^2}{2}+\sum\limits_j c_j\, h_j+\sum\limits_{i,j} w_{ij}\,v_i\, h_j }.\label{eq:lse}
\eeq
If the $m \times n$ matrix, $\vecw$, is stacked to form a vector of dimension $m n$, denoted as $\it{vec}(\vecw)$,
then we can express eq. \eqref{eq:lse} as a function of $\vecw$ and $\vecc$ as,
\beq
g(\theta,\vecv)=  \,\log \sum\limits_{k=1}^{2^{n}} \, \beta_k\,\, e^{\veca_k^T \it{vec}(\vecw)}\,\, e^{\vech^{(k)^T} \vecc}\non
\eeq
where, $\beta_k = e^{-\sum_i \frac{(v_i-b_i)^2}{2}}$, $\veca_k = \it{vec}(\vecv ({\vech^{(k)}})^T)$ and 
$\vech^{(k)},\, k=1,\ldots,2^n$ are all the $n$-bit binary vectors or, equivalently, all possible values of the binary hidden vector $\vech$ (Note that 
$\vecv ({\vech^{(k)}})^T$ is a matrix of dimension  $m \times n$ and $\it{vec}(\vecv ({\vech^{(k)}})^T)$ is a vectorized version of it).
The following Lemma now proves that $g$ is convex \wrt $\vecw$ and $\vecc$.
\begin{lemma}
A function of the following form is convex:
\[
\it{lse}_\veca(\vecu)=\log\sum\limits_{k=1}^N\beta_k \,\,e^{\veca_k^T \vecu}
\]\label{lem_lse}
where, $\veca_1,\ldots,\veca_N$ are some fixed vectors 
in $\Real^d$ and $\beta_k \geq 0, \forall k$.   (Note that the function $\it{lse}_\veca$ maps $\Real^d$ to $\Real$).
\end{lemma}
\noindent
The proof of the above Lemma is given in appendix \ref{a_lse}. 

Hence we can conclude that $g$ is convex \wrt $\vecw,\vecc$.
It is to be noted that the function $g$ is not convex \wrt $\vecb$.
\subsection{Convexity of $f(\theta)$}\label{sub_sec_cvx_f}
As shown in appendix \ref{appendix_f_g}, we can write $f(\theta)$ as,
\beq
f(\theta)= K_f +\log \sum_\vech e^{\sum\limits_j c_j h_j+\sum\limits_i \lb\sum\limits_j w_{ij}\, h_j\rb\lb\frac{\sum\limits_j w_{ij}\, h_j}{2}+b_i\rb}.\non
\eeq
This can also be expressed as,
\beq
f(\theta)= K_f +\,\log \sum\limits_{k=1}^{2^{n}} \, \beta_k e^{s(k)}
\label{eq:lseq}
\eeq
where, $s(k){=}\frac{1}{2}\sum_i (\sum_j w_{ij}\, h_j^k)^2{+}\sum_{i,j} w_{ij}\, h_j^k b_i$ and  $\beta_k {=} e^{\sum_j c_j\,h_j^k}$ with $h^k_j$ being the $j^{th}$ component of the binary 
vector $\vech^k$. It is to be noted that $\beta_k \geq 0, \forall k$. 
Let the $i^{\text{th}}$ row of $\vecw$ be denoted as $\vecw_{i*}$. Then,
\beq
\lb\sum\nolimits_{j} w_{ij}\, h_j^k \rb^2 = \left(\vecw_{i*}^T\, \vech^{K}\right)^2 = \vecw_{i*}^T \,\vech^{K}(\vech^{K})^T \,\vecw_{i*}. \non
\eeq
Using the above and by treating $\it{vec}(\vecw)$ as an $mn$-dimensional vector where rows of $\vecw$ are stacked one below the other (as done earlier), 
the eq. \eqref{eq:lseq} can be rewritten in the following form,
\beq
f(\theta)= K_f +\log\sum\limits_k e^{\vecc^T\vech^{(k)}}\, e^{\frac{1}{2} \it{vec}(\vecw)^T A_k \it{vec}(\vecw)+\veca_k^T \it{vec}(\vecw)}
\label{eq:lseq_vec}
\eeq
Here, $A_k$ is an $ mn \times mn$ block diagonal 
matrix in which each block is a rank one matrix of the form $\vech^k (\vech^{k})^T$ and $\veca_k=\it{vec}(\vecb (\vech^k)^T)$ is a vector of dimension $mn$. 
The  form of $A_k$ ensures that it is a positive semi-definite matrix.
The following Lemma now proves $f$ is convex \wrt $\vecw, \vecc$.
\begin{lemma}
A function of the form 
\[lse_q(\vecu)=\log\sum\limits_i \beta_i e^{\frac{1}{2}\vecu^T A_i \vecu+\veca_i^T \vecu}\] is convex if the matrix $A_i$ is positive semi-definite and $\beta_i \geq 0, \forall i$.\label{lem}
\end{lemma}
\noindent The proof of the above Lemma is given in appendix \ref{lemma_proof}.\footnote{We call the function in Lemma \ref{lem} as a log-sum-exponential-quadratic function and hence, the notation $lse_q$. While the 
convexity of \textit{lse} functions given by \eqref{lse_appendix} in appendix \ref{a_lse} is known, we do not know of any discussion or reported proofs of convexity of the $lse_q$
functions considered here. 
}




Having proved that both $f$ and $g$ are convex \wrt $\vecw$ and $\vecc$, we can now conclude, using eq. \eqref{eq:ll_base}, that the log-likelihood function of the GB-RBM is a 
difference of convex functions \wrt weights, $\vecw$, and bias of hidden units, $\vecc$. 
%
\subsection{Difference of Convex Functions Optimization}\label{subsec_S-DCP}
The  difference of convex programming (DCP) \citep{yuille2002concave,An2005} is a method used for  solving optimization problems of the form,
\beq
\theta^{*}=\arg \min_\theta F(\theta)=\arg \min\limits_\theta \lb f(\theta)-g(\theta)\rb,\non
\eeq
where, both $f$ and $g$ are convex functions of $\theta$. 
In the RBM setting, $F$ corresponds to the negative log-likelihood and the functions $f$ and $g$ correspond to those defined in eq. \eqref{eq:f_g_def}. 
While both $f$ and $g$ are convex and smooth, $F$ is non-convex. 

The DCP is an iterative procedure defined by,
\beq
\theta^{(t+1)}=\arg \min_\theta \lb f(\theta)-\theta^T \nabla g(\theta^{(t)})\rb.\label{eq:CCP_DCA_convx}
\eeq
Note that the optimization problem on the RHS of eq. \eqref{eq:CCP_DCA_convx} is convex and hence,
the DCP consists of solving a convex optimization problem in each iteration. 
It can be shown, using
convexity of $f$ and $g$, that the  iterates defined by eq. \eqref{eq:CCP_DCA_convx} constitute a descent method for minimizing F~\citep{yuille2002concave,An2005}. One can also show that the iterates would constitute a descent on 
$F$ as long as a descent on the objective function in (the RHS of) eq. \eqref{eq:CCP_DCA_convx} is ensured at each iteration even if this optimization problem is not solved exactly.
Such an approach, called the stochastic-DC programming (S-DCP) algorithm, was proposed earlier for learning BB-RBMs \citep{pmlr-v77-upadhya17a}.

In the S-DCP algorithm,  
 the convex optimization problem described in  
eq.\eqref{eq:CCP_DCA_convx} is solved (approximately) using a fixed $d$ number of iterations of stochastic gradient descent (\wrt $\theta$) on $f(\theta)-\theta^T \nabla g(\theta^{(t)},\vecv)$. For this,
the $\nabla f$ is estimated using samples obtained though MCMC, as in the case of CD (the estimate is denoted as $\hat{f'}(\theta,\vecv)$). 
The S-DCP algorithm is described in detail via Algorithm \ref{Alg:S-DCP}.

If the computational cost of one Gibbs transition is $T$ and that of evaluating gradient of $g$ (or $f$) is $L$, the
computational cost of the CD-$K$ algorithm for a mini-batch of size $N_B$ is $N_B(KT+2L)$ whereas
the S-DCP algorithm has cost
$d\, N_B(K' T+L)+N_B L$ when $K'$ MCMC steps and $d$ inner loop iterations are chosen~\citep{pmlr-v77-upadhya17a}.
It may be noted that, by choosing the S-DCP hyperparameters $d$ and $K'$ such that $KT=dK'T+(d-1)L$,  the amount of 
computations required is approximately same as that for CD algorithm with $K$ steps~\citep{pmlr-v77-upadhya17a}. 
If we assume that $(d-1)L$ is small, then 
the computational cost of both CD and S-DCP algorithms can be made approximately equal by choosing $dK'=K$.
However, S-DCP updates the parameter ($\theta$) $d$ times per mini-batch whereas CD 
updates the parameter only once per mini-batch. Therefore, S-DCP computational
cost is slightly more compared to CD (refer table \ref{tab:time_compare} for the comparison of the exact computational time for different algorithms).


In sections \ref{sub_sec_cvx_g} and \ref{sub_sec_cvx_f} we showed that the 
functions $f$ and $g$ are convex only with respect to the parameters $\vecw$ and $\vecc$. Hence, the S-DCP  cannot be directly used for learning the biases of the visible units, $\vecb$.
We could update $\vecb$ outside the S-DCP loop using the same update as in CD.
Instead, we fix this parameter to be the mean of the data directly without training, as suggested in \cite{10.1371/journal.pone.0171015}.
Our experimental results show that even with a fixed $\vecb$, we could still learn good models.

With regard to the variance parameter, the choice of 
$\sigma_i^2=1, \forall i$, indicates that we learn a Gaussian with unit variance as the conditional distribution of visible units.
Therefore, we suitably normalize the input data to ensure unit variance and compare the performance of the S-DCP algorithm with that of the CD and
PCD algorithms.
We also explore the performance of these algorithms when the variance parameter not fixed, but is learnt by the model.
In this case, only $\vecw$ and $\vecc$ are updated in the inner loop of the S-DCP algorithm and the variance parameter
is updated only once per mini-batch outside the S-DCP loop. We use the re-parameterization of the variance parameter as in \cite{cho}
to constrain them to stay positive during the update.
A detailed description of this variant of the S-DCP algorithm is given in  
Algorithm \ref{Alg:S-DCP-Var}.

\begin{algorithm}[tb]
   \caption{S-DCP update for a single training sample $\vecv$ \citep{pmlr-v77-upadhya17a}. The estimate of the $\nabla f$ is denoted as $\hat{f'}(\theta,\vecv)$.}\label{Alg:S-DCP}
\begin{algorithmic}
   \STATE {\bfseries Input:} $\vecv,\theta^{(t)}=\{\vecw^{(t)},\vecc^{(t)}\},\eta,d, K'$
   \STATE Initialize $\sth^{(0)}=\theta^{(t)},\sv^{(0)}=\vecv$
       \FOR{$l=0$ {\bfseries to}  $d-1$}
           \FOR{$k=0$ {\bfseries to}  $K'-1$}
               \STATE sample $h_i^{(k)}\sim p(h_i\vert\sv^{(k)},\sth^{(l)}), \forall i$
	       \STATE sample $\tilde{v}_j^{(k+1)}\sim p(v_j\vert\vech^{(k)},\sth^{(l)}), \forall j$
           \ENDFOR
           \STATE $\sth^{(l+1)}=\sth^{(l)}-\eta \ls\hat{f'}(\sth^{(l)},\sv^{(K')})-\nabla g (\theta^{(t)},\vecv)\rsq $  
           \STATE $\sv^{(0)}=\sv^{(K')}$
   \ENDFOR 
\STATE {\bfseries Output:}  $\theta^{(t+1)}=\sth^{(d)}$
   \end{algorithmic}
\end{algorithm}

\begin{algorithm}[tb]
   \caption{S-DCP update for a single training sample $\vecv$ when learning variance}\label{Alg:S-DCP-Var}
\begin{algorithmic}
   \STATE {\bfseries Input:} $\vecv,\theta^{(t)}=\{\vecw^{(t)},\vecc^{(t)}\},\bm{\sigma}^{(t)},\eta,d, K'$
   \STATE Initialize $\sth^{(0)}=\theta^{(t)},\sv^{(0)}=\vecv$
       \FOR{$l=0$ {\bfseries to}  $d-1$}
           \FOR{$k=0$ {\bfseries to}  $K'-1$}
               \STATE sample $h_i^{(k)}\sim p(h_i\vert\sv^{(k)},\sth^{(l)}), \forall i$
	       \STATE sample $\tilde{v}_j^{(k+1)}\sim p(v_j\vert\vech^{(k)},\sth^{(l)}), \forall j$
           \ENDFOR
           \STATE $\sth^{(l+1)}=\sth^{(l)}-\eta \ls\hat{f'}(\sth^{(l)},\sv^{(K')})-\nabla g (\theta^{(t)},\vecv)\rsq $  
           \STATE $\sv^{(0)}=\sv^{(K')}$
   \ENDFOR 
\STATE Update $\bm{\sigma}$ using re-parameterization as in \cite{cho}. 
\STATE {\bfseries Output:}  $\theta^{(t+1)}=\sth^{(d)}, \bm{\sigma}^{(t+1)}$
   \end{algorithmic}
\end{algorithm}

\section{Experiments and Discussion}\label{sec:experiments}
\subsection{Datasets}
In order to analyse the performance of the different algorithms, we consider four benchmark grey scale image datasets namely 
Natural images  \citep{RIS_0}, Olivetti faces \citep{341300}, MNIST \citep{LeCun:1998} and Fashion-MNIST (FMNIST)\citep{xiao2017/online}.
The details of each of these datasets, the data dimension ($m$), number of training
samples ($N_{tr}$) and number of test samples ($N_{te}$) along with the other hyperparameters of algorithms (to be explained in Section \ref{subsec:Exp_setup}) are 
provided in Table \ref{tab:dataset_info}.
\setlength{\tabcolsep}{0.37em}
\begin{table}[ht!]                                                                                          
\centering                             
\caption{The Details of the Datasets and the Hyperparameters}\label{tab:dataset_info}
\begin{tabular}{|c|c|c|c|c|c|c|c|c|c|}
\hline
    Dataset & $m$ & $N_{tr}$  & $N_{te}$ & $n$ & $N_E$ & $N_B$  & $K$ & $d$ & $K'$ \\
    \hline
    Nat. Images& $196$ & $40K$ & $30K$ & $196$ & $200$ & $100$ & $12$ & $3$ & $4$\\
    \hline
MNIST & $784$ & $60K$ & $10K$ & $200$ & $300$ & $200$ & $24$ & $4$ & $6$\\
    \hline
FMNIST & $784$ & $60K$ & $10K$ & $200$ & $300$ & $200$ & $24$ & $4$ & $6$\\
    \hline
Olivetti Faces & $4096$ & $400$ & $-$ & $500$ & $3000$ & $10$ & $12$ & $3$ & $4$\\
    \hline
    \end{tabular}
\end{table}

Similar to \cite{DBLP:journals/corr/WangMW14} the Natural images dataset used is a subset of the Van Hateren’s Natural image database \citep{RIS_0}. 
These images are whitened using Zero-phase Component Analysis (ZCA).  
As mentioned in section \ref{sec:background}, the features learnt using GB-RBM on this dataset are comparable to those learnt using sparse coding and Independent Component Analysis (ICA) \citep{DBLP:journals/corr/WangMW14}.

In addition to the Natural image dataset, we also consider the Olivetti faces dataset which was used to study deep 
Boltzmann machines \citep{6706831}.
There are ten different images for each of the $40$ subjects in this dataset. 
They are whitened using ZCA.
All the images are used to train the GB-RBM model primarily because the number of data samples are less in this dataset.
Training using all the images also alleviates the sensitivity issues of the RBM (RBMs are sensitive to
translation in images) which could be caused by the mismatch in the alignment of training and testing images.

The MNIST dataset contains handwritten digits, whereas the FMNIST contains images of fashion products.   
All these images are normalized to have mean $0$ and variance $1$. 


\subsection{Experimental setup}\label{subsec:Exp_setup}
For each dataset, the hyperparameters such as the number of hidden nodes ($n$), mini-batch size ($N_B$) etc are chosen to be the same across all the algorithms, as provided in Table \ref{tab:dataset_info}. The learning rate ($\eta$) is chosen 
separately  for each algorithm as the maximum possible value (such that, increasing it further will cause divergence of the log-likelihood).
The number of hidden nodes (denoted by $n$) in the RBM models for different datasets is shown in Table~\ref{tab:dataset_info}. 
To study the speed of learning, we train the GB-RBM model for a fixed number
of epochs (denoted as $N_E$ in Table~\ref{tab:dataset_info}).
In each epoch, the mini-batch learning procedure is employed and the training data is shuffled after every epoch.
In order to get an unbiased comparison, we did not use momentum or weight decay in any of the algorithms.


We evaluate the algorithms using the performance measures obtained from multiple trials, where each trial 
fixes the initial configuration of the weights and biases as follows. 
The weights are initialized to samples drawn from a uniform distribution with support $[-a,a]$ where, $a=6/\sqrt{m+n}$.
The bias of the hidden units is initialized to $c_j=-\frac{\| b-W_{*j} \|^2-\|b\|^2}{2}+\log \tau_j$, as proposed in \cite{DBLP:journals/corr/WangMW14} where,
$\tau_j$ is fixed to $0.01$. 
The bias of the visible units ($\vecb$) is not learnt (as explained in the section \ref{subsec_S-DCP}) but fixed to 
the mean of the training samples. 


We compare the performance of S-DCP with CD and PCD keeping the computational complexity of S-DCP roughly the same as that of CD/PCD by  
choosing $K,d$ and $K'$ such that $K=dK'$~\citep{pmlr-v77-upadhya17a}.
We have experimentally observed that a large $K$ generally results in a sensible generative model, also supported by the findings from \cite{DBLP:journals/corr/WangMW14, carlson2015stochastic}. Therefore, we 
choose $K=24$ in CD and PCD (with $d=4,K'=6$ for S-DCP) for learning 
MNIST and FMNIST datasets and we choose $K=12$ in CD and PCD (with $d=3,K'=4$ for S-DCP) for the other two datasets, as shown in Table \ref{tab:dataset_info}.

The algorithms  are implemented using Python and CUDAMat (a CUDA-based matrix class for Python bindings) \citep{Mnih2009CUDAMatAC} on a system with Intel 
processor $i7-7700$ ($4$ CPU cores and processor base frequency $3.60$ GHz), NVIDIA Titan X Pascal GPU
and a $16$ GB RAM configuration.


\subsection{Evaluation Criterion}\label{subsec:Eval}
We compare the performance of different algorithms based on the following criteria:
speed of learning, quality and generalization capacity of the learnt model,
quality of the learnt filters and generative ability of the learnt model.
 
We use the achieved log-likelihood on the train and the test samples to evaluate the quality of learnt models \citep{hinton2010practical}.
While the achieved average train log-likelihood provides a reasonable measure to compare 
the quality of the learnt model, its evolution (over epochs) indicates the speed 
of learning.
The average train log-likelihood (denoted as 
ATLL) is evaluated as, 
\beq
ATLL= \frac{1}{N}\sum_{i=1}^N \log \, p(\vecv_{train}^{(i)}\vert\theta)
\eeq
where, $\vecv_{train}^{(i)}$ denotes the $i^{\text{th}}$ training sample.
The average test log-likelihood (denoted as ATeLL), on the other hand, provides a good indication of the generalization 
capacity of the model. Similar to ATLL, we evaluate the average test log-likelihood by using the test samples rather than the training samples.
The computation of ATLL and ATeLL require the estimate of the partition function. It is  
estimated using annealed importance sampling \citep{neal2001annealed}
with $100$ particles and $10000$ intermediate distributions according to a linear temperature scale between $0$ and $1$. 
In the following section, we show the evolution (over epochs) of 
mean and maximum (over all trials) of the average train and 
test log-likelihood, for all the algorithms considered, for both the cases of fixed variance and learnt variance.

The learning procedure is effective when the weights learnt by the model are
able to capture the important/crucial features of the dataset. 
%
Therefore, we interpret the model weights as filters and examine the filters learnt. 
The evaluation in terms of the generative ability of the 
learnt models is carried out by observing the samples that 
these models generate.
For this, we randomly initialize the states of the
visible units (in the learnt model) and run the alternating Gibbs 
Sampler for 200 steps and plot the state of the visible units.

\subsection{Performance Comparison}
\subsubsection{Speed of Learning}\label{susub:speed}
As mentioned in section \ref{subsec:Exp_setup} all the three algorithms have comparable computational load (per mini-batch) if  we select $K=d\,K'$.  
However, the computations performed by the different algorithms are not identical (S-DCP updates the parameter
$d$ times per mini-batch whereas CD updates
the parameter only once per mini-batch). Therefore,
we present the actual computational time of different algorithms. 
Table \ref{tab:time_compare} provides the mean and the standard deviation $(\sigma_t)$ of the utilized time (in seconds), computed  over $20$ trials, for a fixed $100$ epochs of learning 
by each of the algorithms for the different datasets considered.
As seen from the table, the computation time per epoch is roughly the same for all algorithms.
 \begin{table}[ht!]                                                                                          
\centering                             
\caption{The system time statistics (The mean and standard deviation of the system time in seconds) for different datasets}\label{tab:time_compare}
\begin{tabular}{|c|cc|cc|cc|}
\hline
    Algorithm & \multicolumn{2}{|r}{Natural Images} & \multicolumn{2}{|r}{Olivetti Faces} & \multicolumn{2}{|r|}{MNIST/FMNIST}\\
    & Mean & $\sigma_t$
    & Mean & $\sigma_t$
    & Mean & $\sigma_t$   \\
    \hline
    CD & 40.42  & 0.29 & 7.96 & 0.03 & 77.94 &  0.19\\
 \hline
     PCD & 41.74  &  0.26 & 8.03 & 0.03  & 79.05 & 0.15\\
 \hline
 S-DCP &  43.87 &  0.23 & 9.19  & 0.02 & 88.64  & 0.11\\
 \hline
\end{tabular}
\end{table}

\begin{figure}
\centering
 \includegraphics[width=0.95\columnwidth]{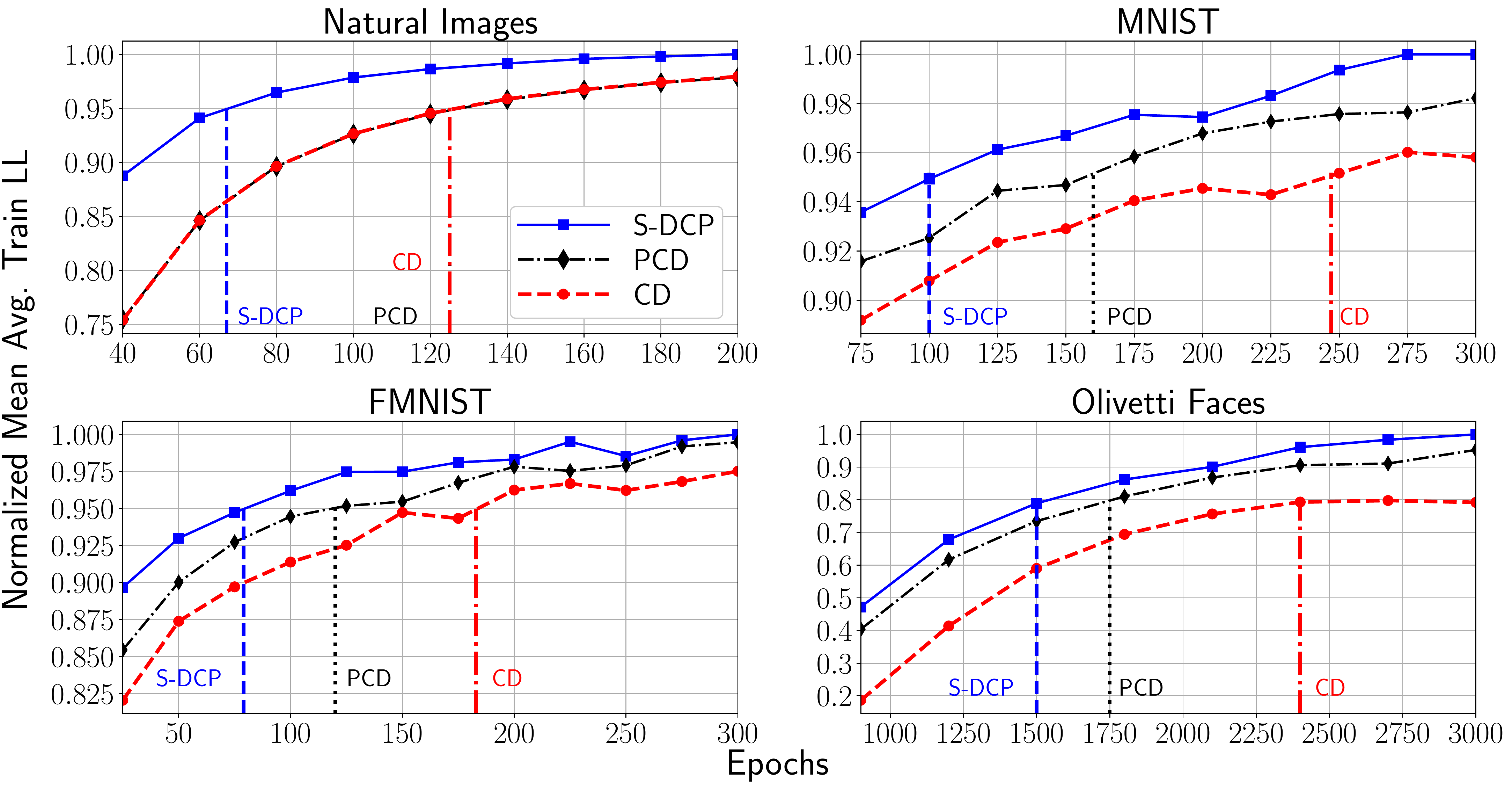}
\caption{The normalized mean average train log-likelihood achieved with different algorithms on all the datasets with fixed variance. For S-DCP, Algorithm \ref{Alg:S-DCP} is used.}\label{Fig:ll_norm_atll_compare}
\end{figure}

To compare the speed of learning as well as the quality of the learnt models, we plot the progress of the {\em normalized} mean average train log-likelihood over epochs for different algorithms in Fig.~\ref{Fig:ll_norm_atll_compare}.
The log-likelihood is normalized using min-max normalization to restrict the range between 0 and 1, and to easily compare the learning speeds.
For each dataset we take the maximum and minimum log-likelihoods among all
algorithms to do this normalization.
From Fig.~\ref{Fig:ll_norm_atll_compare}, we observe that
the log-likelihood achieved at the end of training is different for each algorithm.
The average train log-likelihood achieved
by the S-DCP algorithm is superior to those
achieved by the CD and PCD algorithms across all datasets.

In order to better illustrate the efficiency of the S-DCP algorithm, we choose a reference log-likelihood (denoted as Ref. LL), as shown in Fig.~\ref{Fig:ll_norm_atll_compare}.
For the Natural image,
MNIST and FMNIST datasets the Ref. LL is chosen as 
$95\%$ (i.e., 95\% of the highest ATLL achieved).
For the Olivetti faces
dataset we choose the Ref. LL as the maximum log-likelihood achieved
by the CD algorithm (since CD attains the least log-likelihood among all the algorithms) which is approximately
$80\%$ of the maximum.
Table~\ref{tab:time_epoch_compare} provides the Ref. LL along with the number of epochs and the computational
time required to achieve this for the different algorithms.

Both, Table~\ref{tab:time_epoch_compare} and Fig.~\ref{Fig:ll_norm_atll_compare}, demonstrate that the S-DCP algorithm is much faster compared to other algorithms both in terms of the number of epochs and the actual computational time.
For instance, the speed of learning on the Natural image dataset, indicated in the Fig.~\ref{Fig:ll_norm_atll_compare} by the epoch at which the model achieves $95\%$
of the highest mean ATLL, shows that CD and PCD algorithms require around $60$ more epochs of training compared to the
S-DCP algorithm which translates to approximately twice the required computation time. 
On MNIST and FMNIST datasets, S-DCP is approximately twice as fast as CD and $1.4$ times as fast as the PCD. 
On Olivetti faces dataset, the CD algorithm achieves only $80\%$ of the final log-likelihood achieved by the S-DCP algorithm 
and S-DCP achieves this value approximately $900$ and $250$ epochs earlier than CD and PCD, respectively. 
In terms of the computational time (to learn the Olivetti faces dataset) S-DCP takes about $25\%$ less time compared to CD though the times taken by PCD and S-DCP are about the same.  

Thus, we find that S-DCP has a higher speed of learning compared to the other algorithms. We also observed a similar behaviour when the normalized maximum ATLL is considered instead of normalized mean ATLL.

\setlength{\tabcolsep}{0.37em}
\begin{table}
\centering                             
\caption{The number of epochs and system time statistics (in seconds) for achieving the reference normalized mean average log-likelihood by different algorithms for each datasets.}\label{tab:time_epoch_compare}
\scriptsize{
  \begin{tabular}{|c|c|r|r|r|r|r|r|}
    \hline
    {Dataset} & {Ref. LL}  & \multicolumn{2}{c|} {CD} & \multicolumn{2}{c|} {PCD} & \multicolumn{2}{c|} {S-DCP} \\ \cline{3-8}
    & {($\%$)} & {Epoch} & \textit{Time} & Epoch & \textit{Time} & Epoch & \textit{Time} \\
    \hline
    Nat. Images &  $95$ & $125$ & $\mathit{50.52}$ & $125$ & $\mathit{52.17}$ & $60$ & $\bm{\mathit{29.39}}$\\
    \hline
    MNIST &  $95$ & $247$ & $\mathit{192.51}$ & $160$ & $\mathit{126.48}$ & $100$ & $\bm{\mathit{88.64}}$\\
    \hline
    FMNIST &  $95$ & $183$  & $\mathit{142.63}$ & $120$ & $\mathit{94.86}$ & $79$  & $\bm{\mathit{70.02}}$\\
    \hline 
Olivetti Faces &  $80$ & $2400$  & $\mathit{191.04}$ & $1750$  & $\mathit{140.52}$ & $1500$  & \bm{$\mathit{137.85}}$\\
    \hline   
    \end{tabular}
              }
\end{table}

\subsubsection{Quality and Generalization capacity}
As mentioned earlier, the quality of the trained model could be captured by the average train log-likelihood (ATLL) and the generalization capacity could be inferred from the average test log-likelihood (ATeLL). We 
now show these results in both the cases of fixed variance and learnt variance.

\begin{figure}
\centering
 \includegraphics[width=0.49\textwidth]{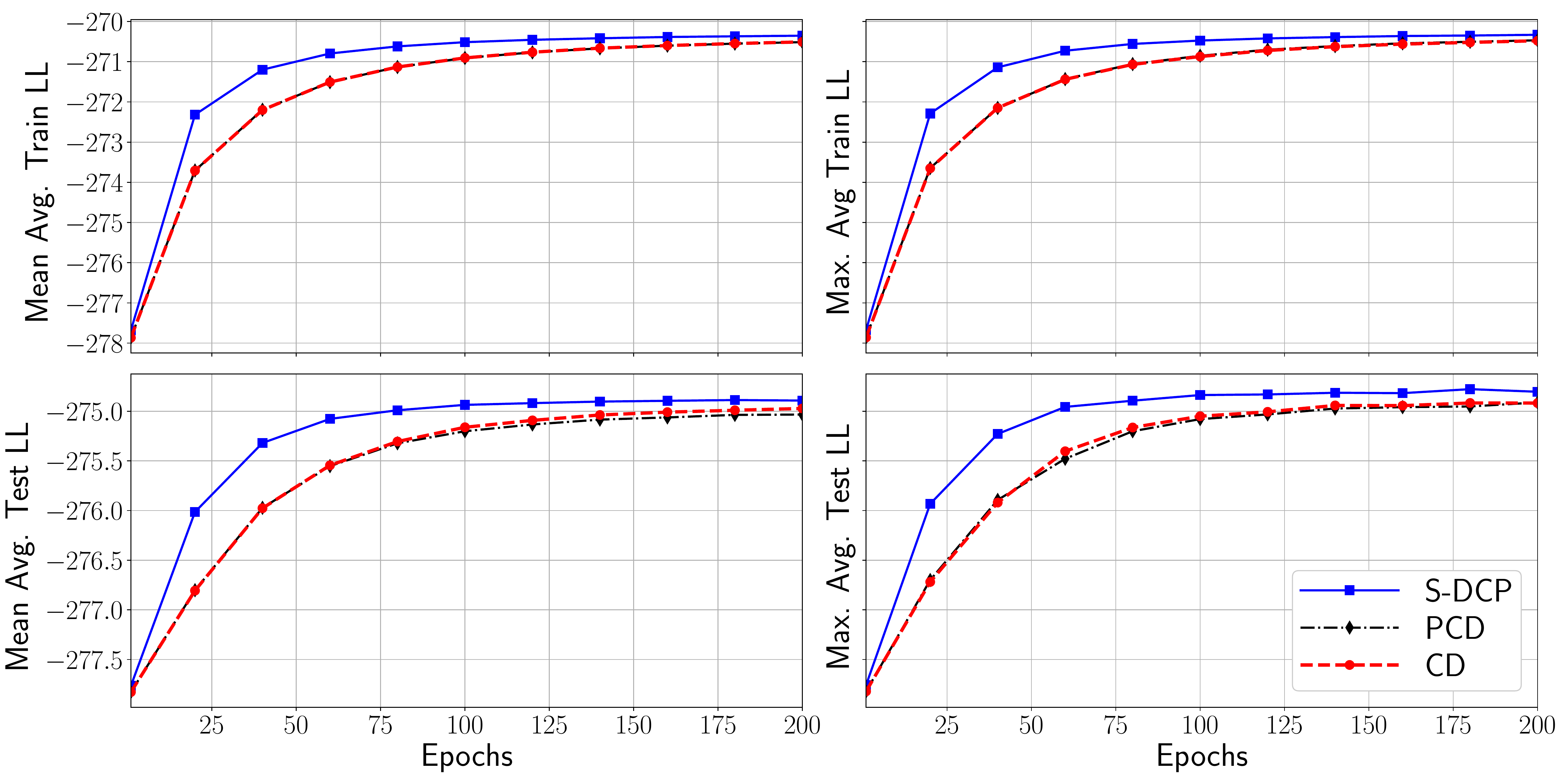}
  \includegraphics[width=0.49\textwidth]{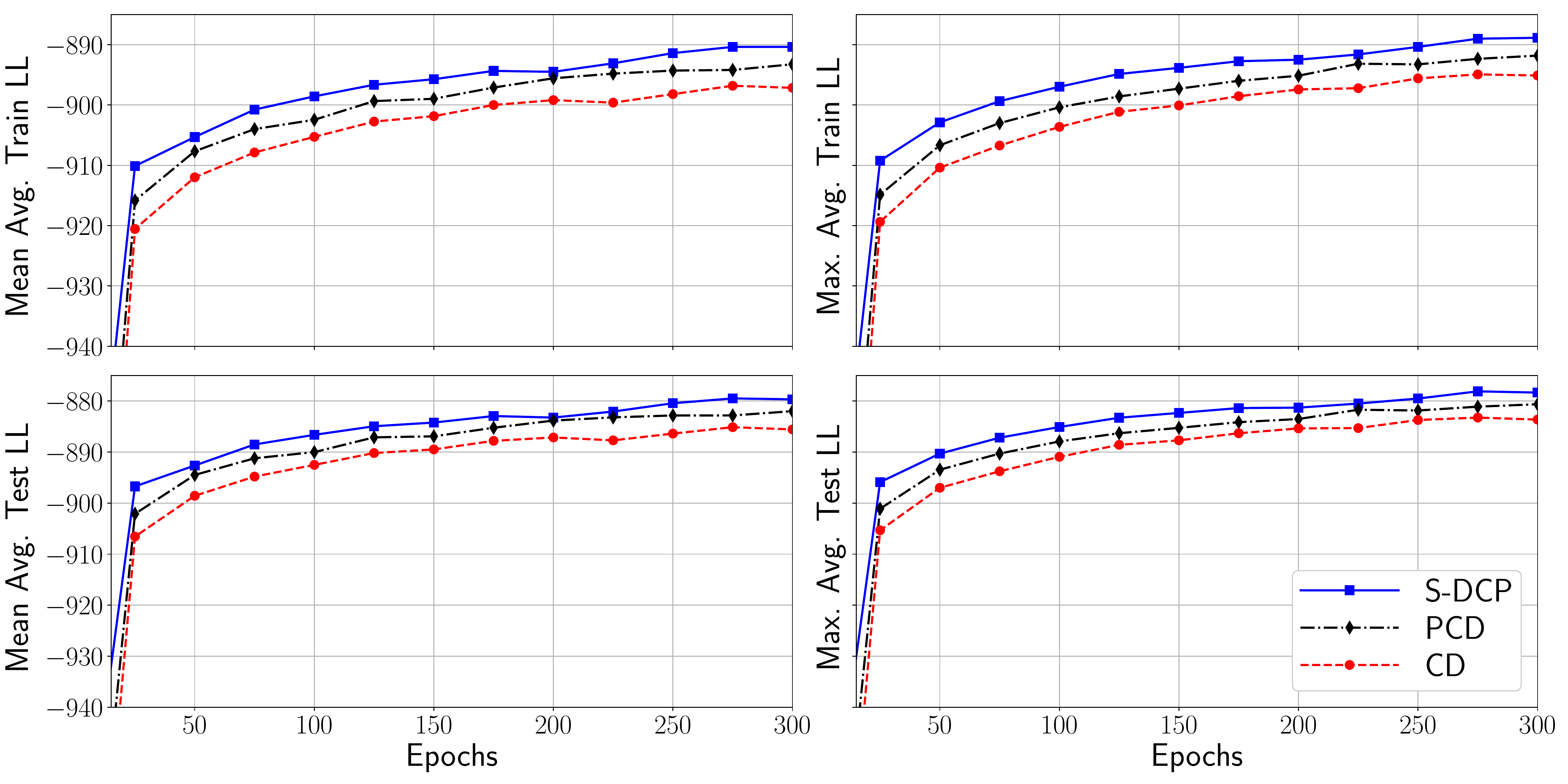}
 \includegraphics[width=0.49\textwidth]{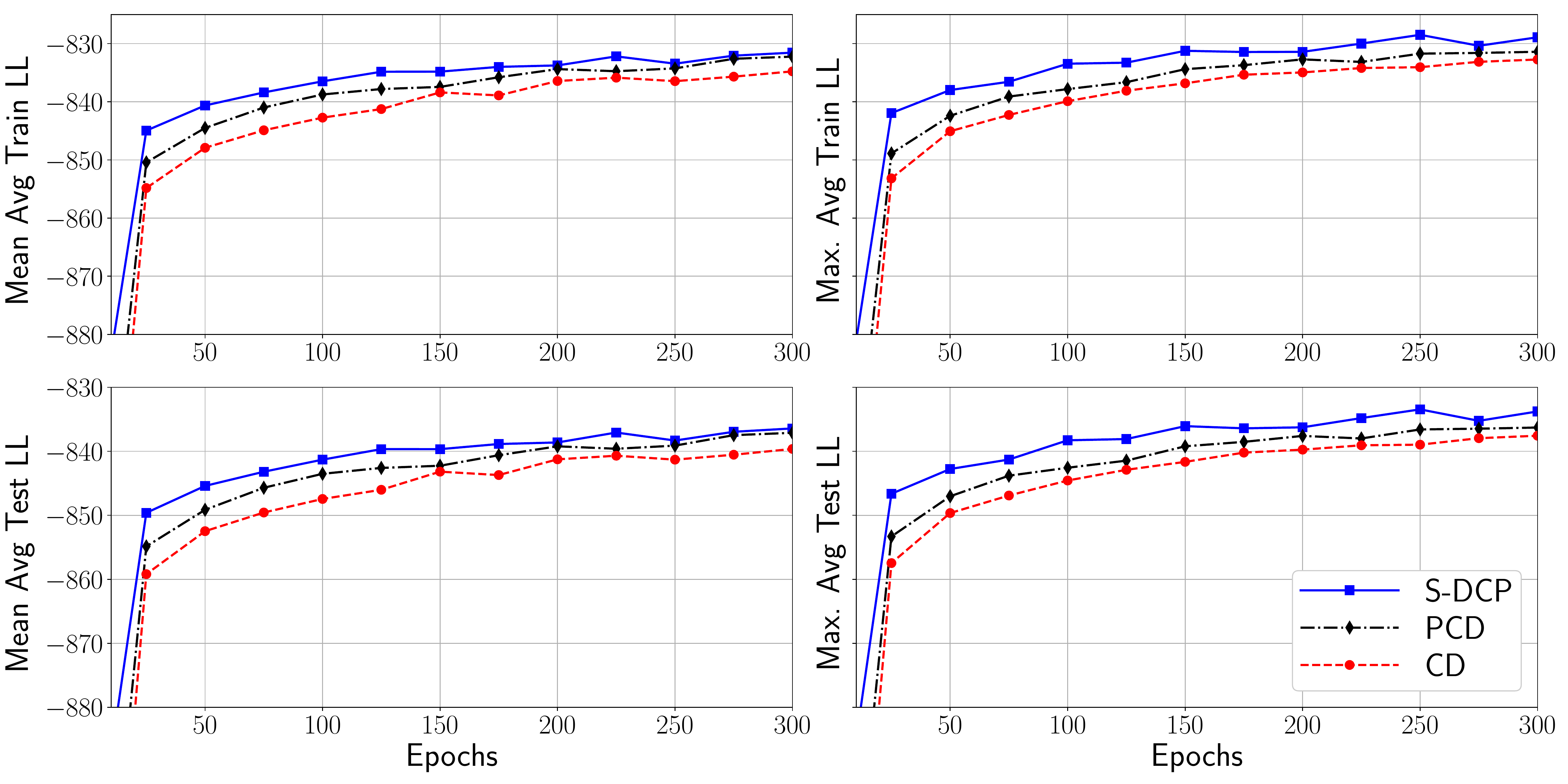}
 \includegraphics[width=0.49\textwidth]{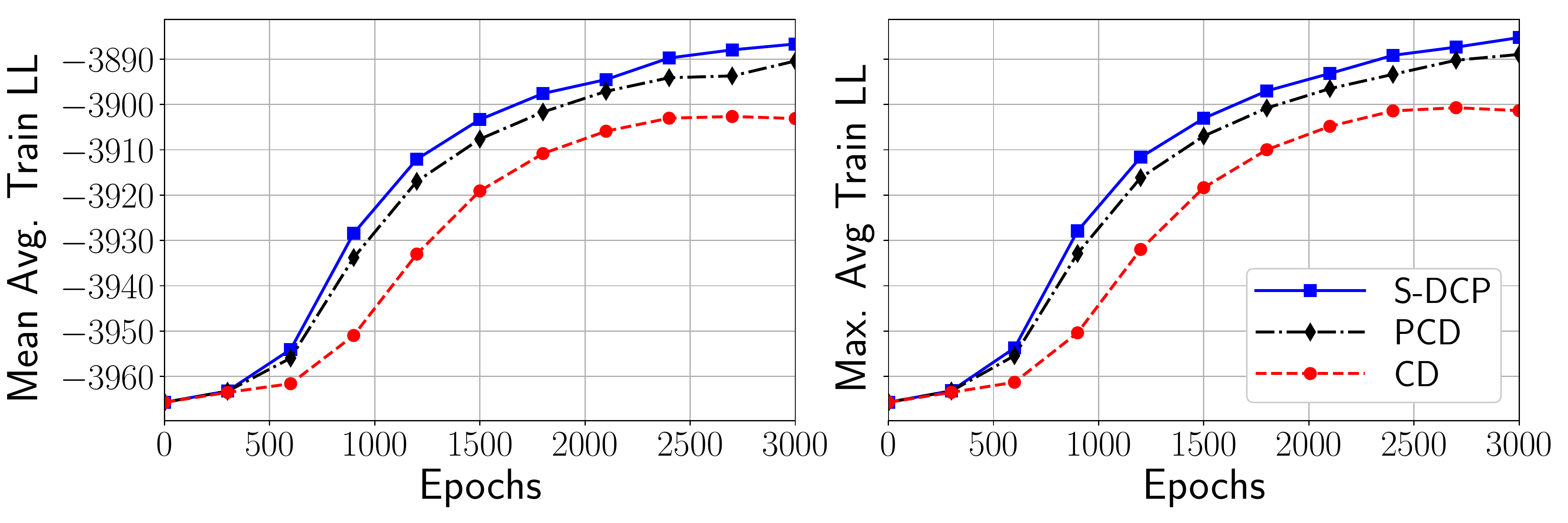}  
 \caption{The performance of different algorithms on each dataset with fixed variance. For S-DCP, Algorithm \ref{Alg:S-DCP} is used. (Note that, unlike in Fig.~\ref{Fig:ll_norm_atll_compare}, here we show the evolution of log likelihood starting from the first epoch; consequently here the range of $y$-axis is more and hence the scale on $y$-axis is compressed).}\label{fig:ll_var}
\end{figure}

\noindent\textbf{Fixed Variance:}
The evolution of the train and the test log-likelihoods shown in Fig. \ref{fig:ll_var}, for each dataset,
demonstrates that the S-DCP algorithm performs consistently better compared to the other two algorithms across all the datasets.

\begin{figure}
\centering
 \includegraphics[width=0.49\textwidth]{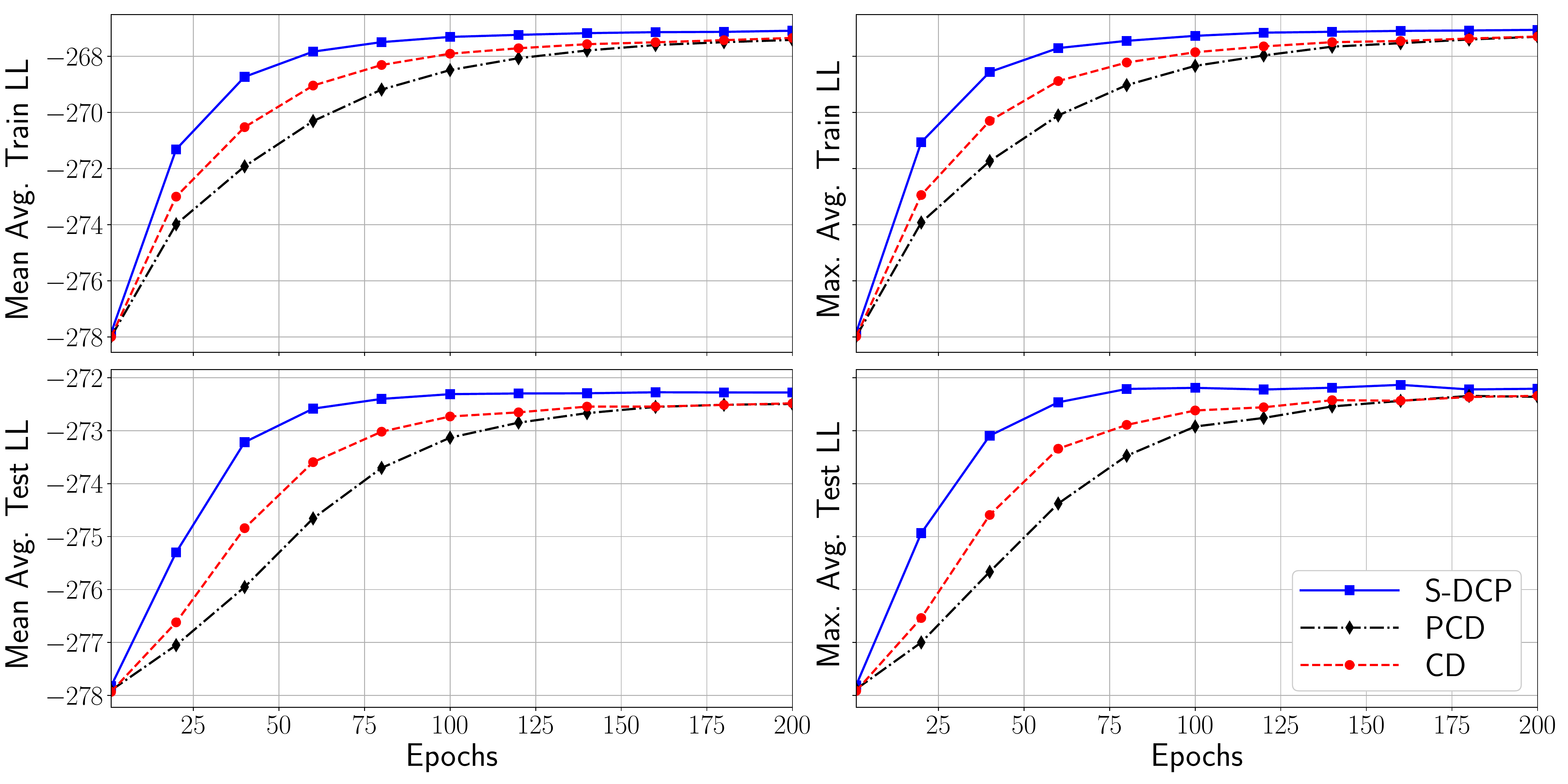}
\includegraphics[width=0.49\textwidth]{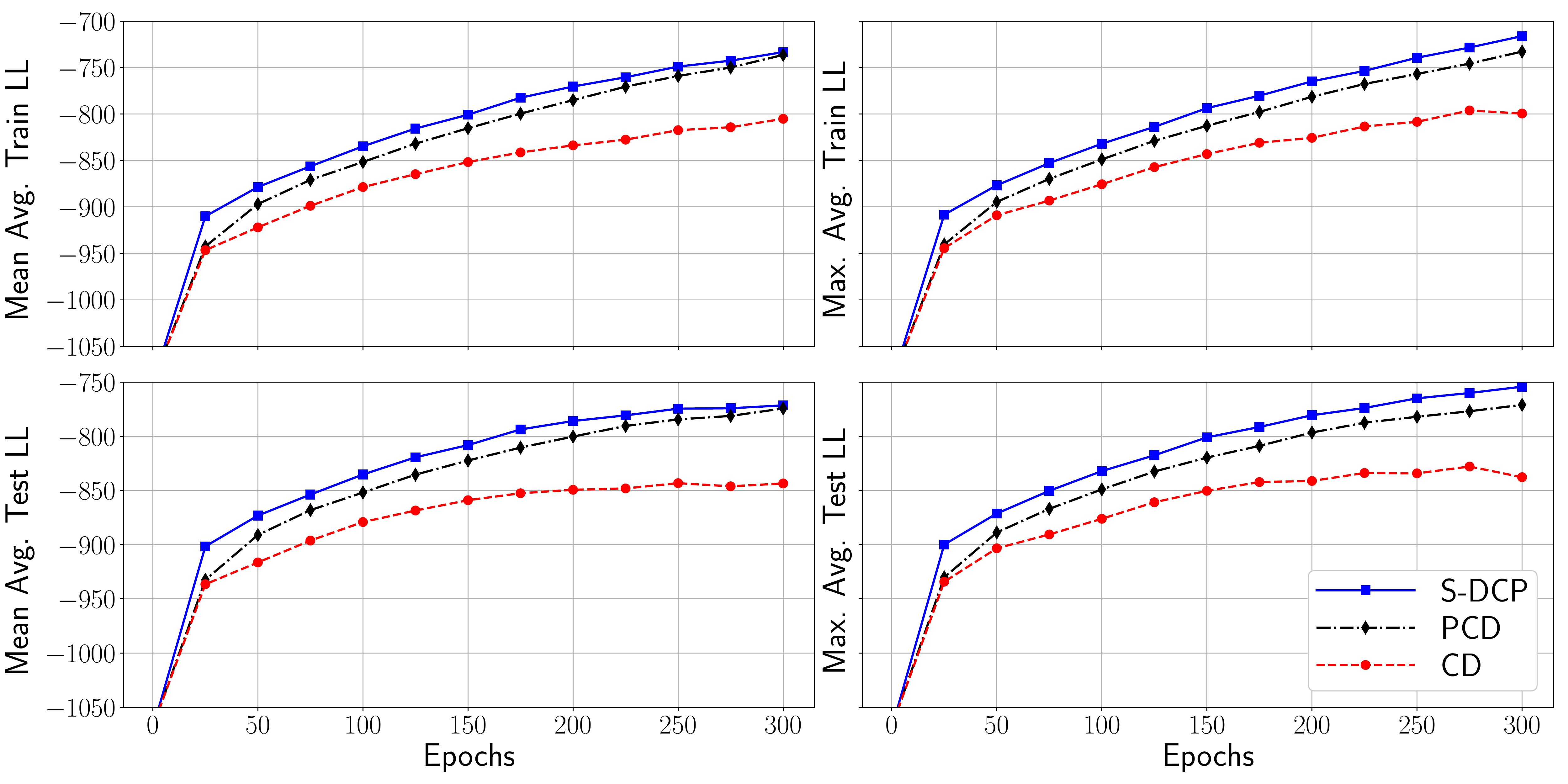}
\includegraphics[width=0.49\textwidth]{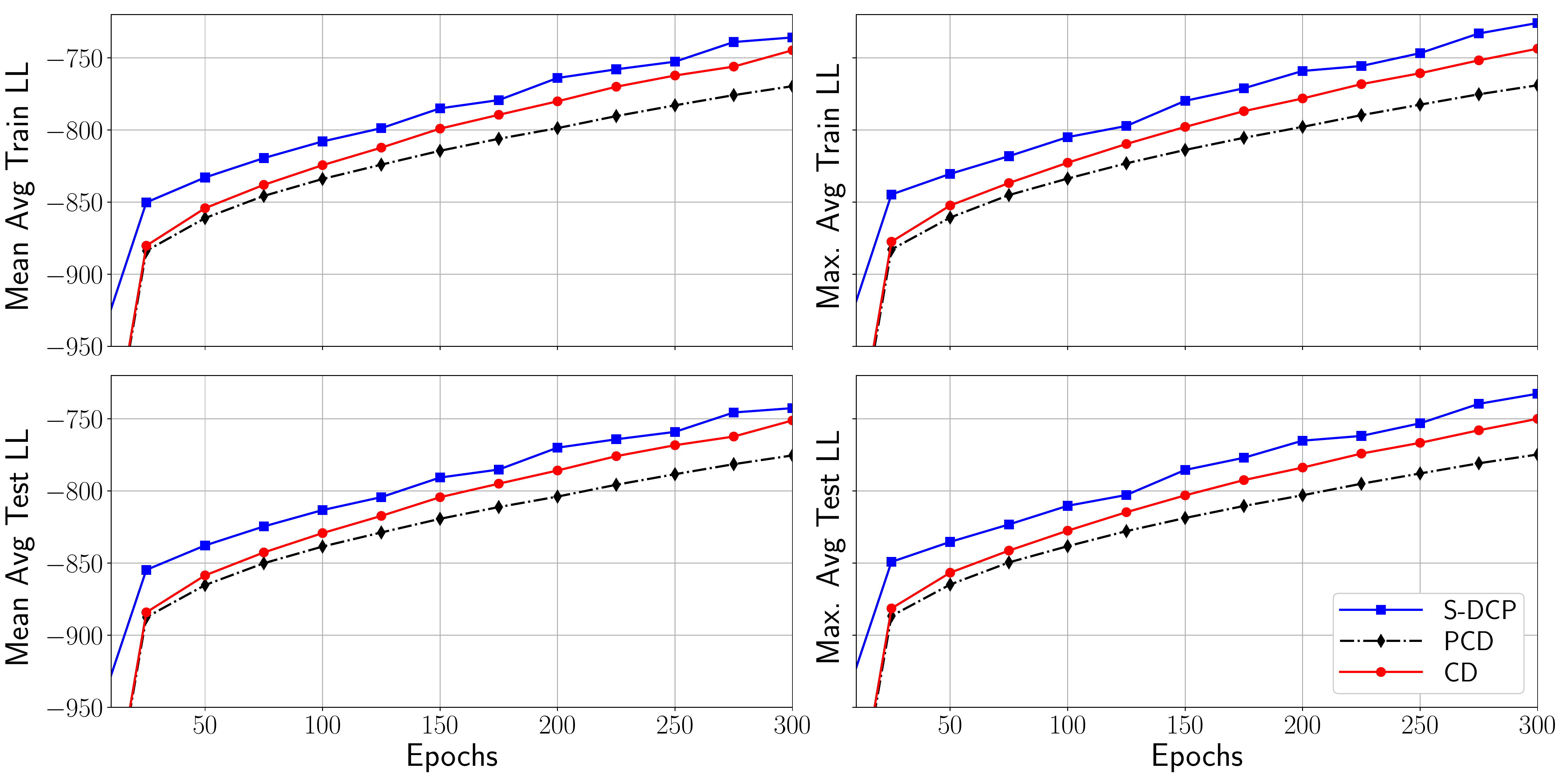}
  \includegraphics[width=0.49\textwidth]{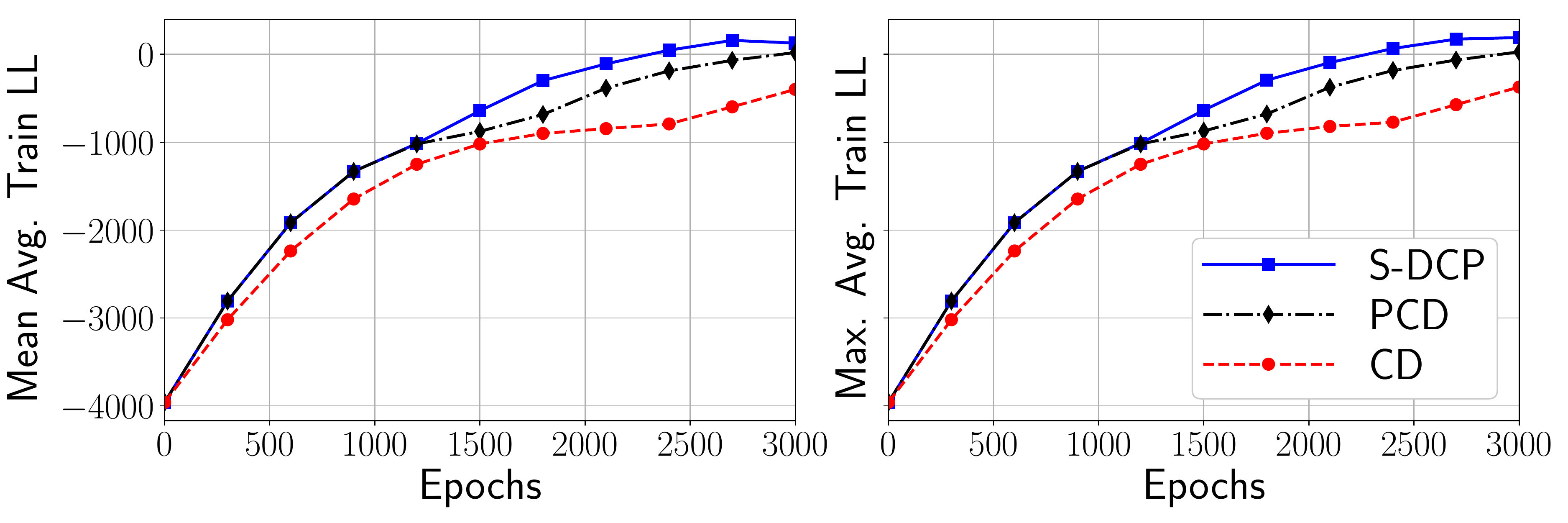}  
\caption{The performance of different algorithms on each dataset  while also learning variance. For S-DCP, Algorithm \ref{Alg:S-DCP-Var} is used. It may be noted that the average train log-likelihood for the Olivetti faces dataset evolves to reach positive value since the learnt variance is small and contributes high negative value
to the log partition function.}\label{fig:ll_diff_var}
\end{figure}

%
\noindent\textbf{Learning Variance:}
 The evolution of the train and the test log-likelihoods, over epochs, when the variance parameter is also learnt using the Algorithm \ref{Alg:S-DCP-Var}, is shown in Fig. \ref{fig:ll_diff_var} for different datasets.
We observe that in this case also S-DCP outperforms both the CD and PCD algorithms. 


%
The higher train and test log-likelihood (across datasets) achieved by the S-DCP compared to those achieved using CD and PCD  indicates the better quality and generalization capacity 
of the models learnt using  the S-DCP algorithm.

\begin{figure}
\centering
  \includegraphics[width=0.35\textwidth]{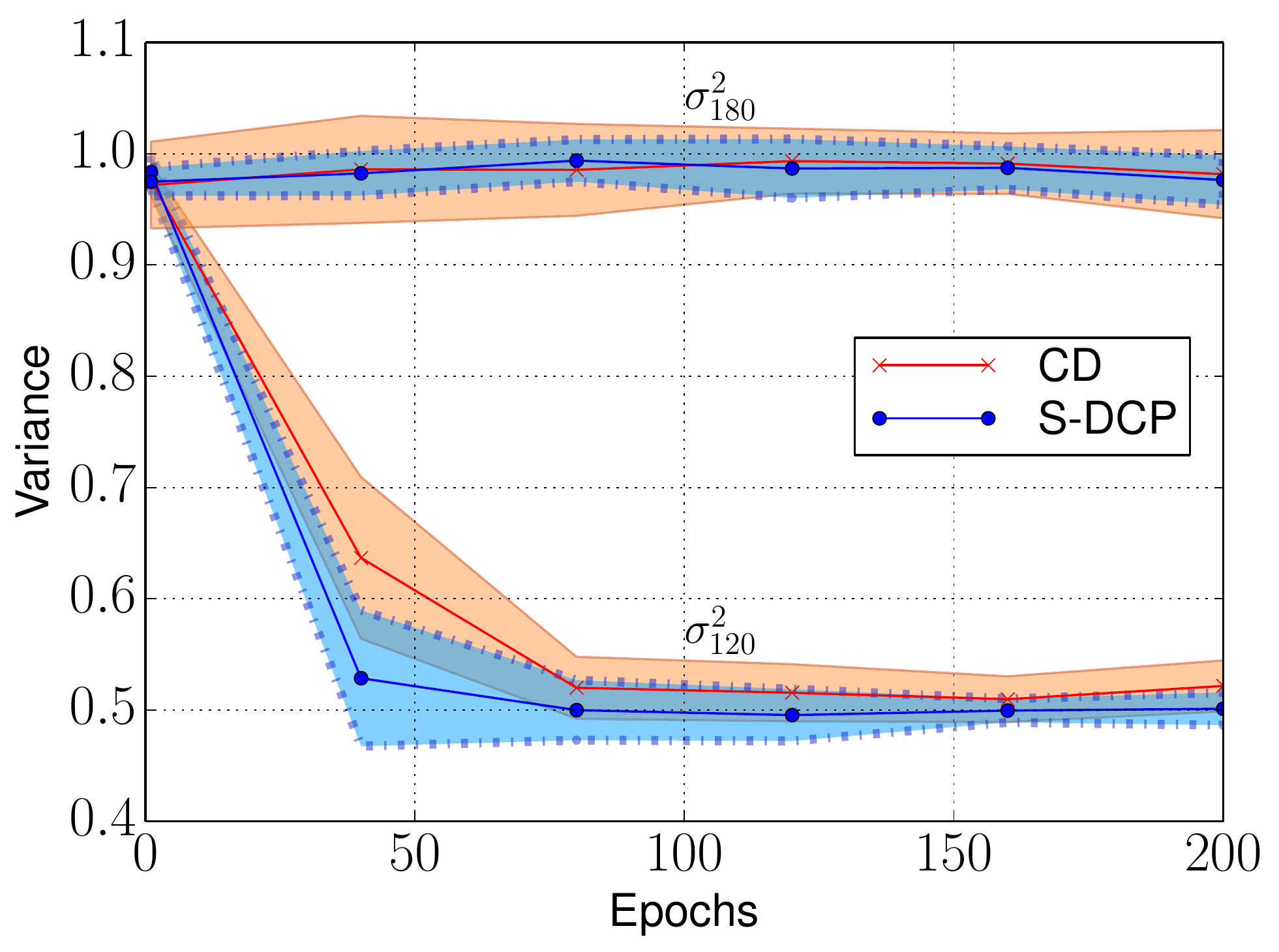}
 \includegraphics[width=0.35\textwidth]{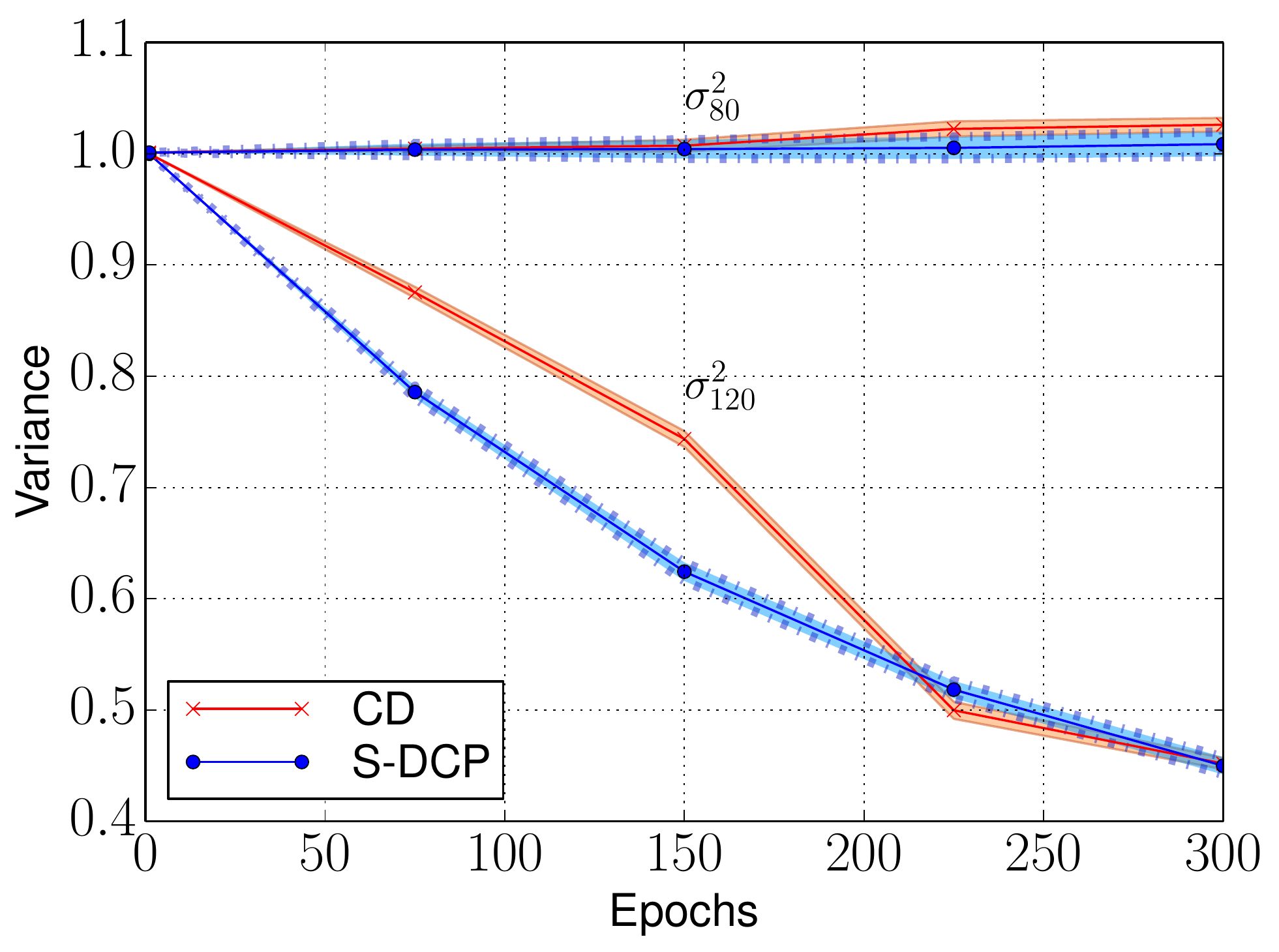}  
  \caption{The evolution of the variance of two specific visible units as learning progresses by CD and S-DCP algorithms on Natural images and MNIST dataset.
For S-DCP, Algorithm 2 is used.}\label{fig:ll_var_evolve}
\end{figure}

In Fig. \ref{fig:ll_var_evolve}, we plot the typical evolution of the variance as 
learning progresses for Natural images and MNIST dataset using CD and S-DCP algorithm (We have observed that PCD learns similar models as that of CD
in terms of learnt variance and hence to avoid clutter PCD plots are omitted). We selected two visible units, and plot the mean (over $20$ trials) of 
the variances of these 
two units as learning evolves. We also show the variance over $20$ trials of this evolution as a shaded region around the plot 
of the mean. These two units seem to be typical of a large number of visible units. All units start with unit variance and most units 
either settle to a variance close to $1$ or much lower.  

From the Fig. \ref{fig:ll_var_evolve}, we observe that both CD and S-DCP converge to 
approximately similar variance values though S-DCP converges faster. We have observed 
similar behaviour with the other two datasets. 


\subsubsection{Quality of the learnt filters}
We next look at the filters learnt by different algorithms. 
In the following discussion, we present the results only for the fixed variance case. The same conclusions hold true even in the case of  
learning the variance.

\begin{figure}
\centering
 \includegraphics[width=0.59\columnwidth]{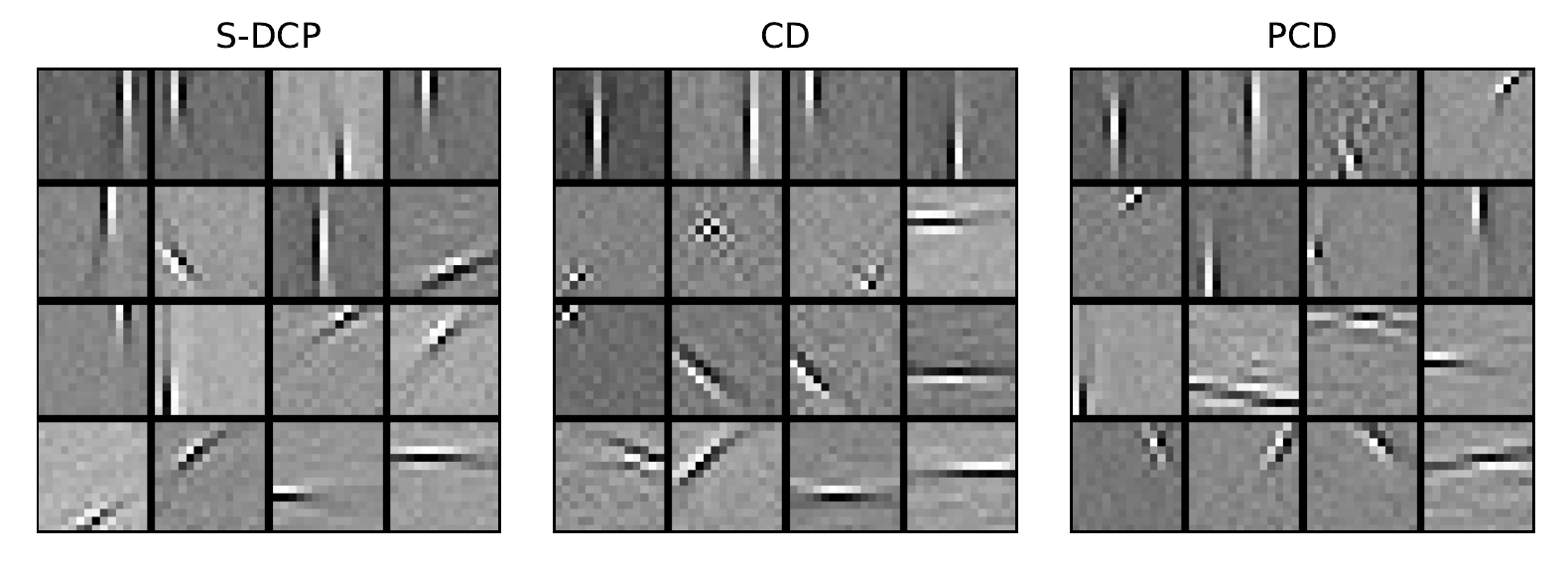}\label{fig:weights_natimages}
  \includegraphics[width=0.59\columnwidth]{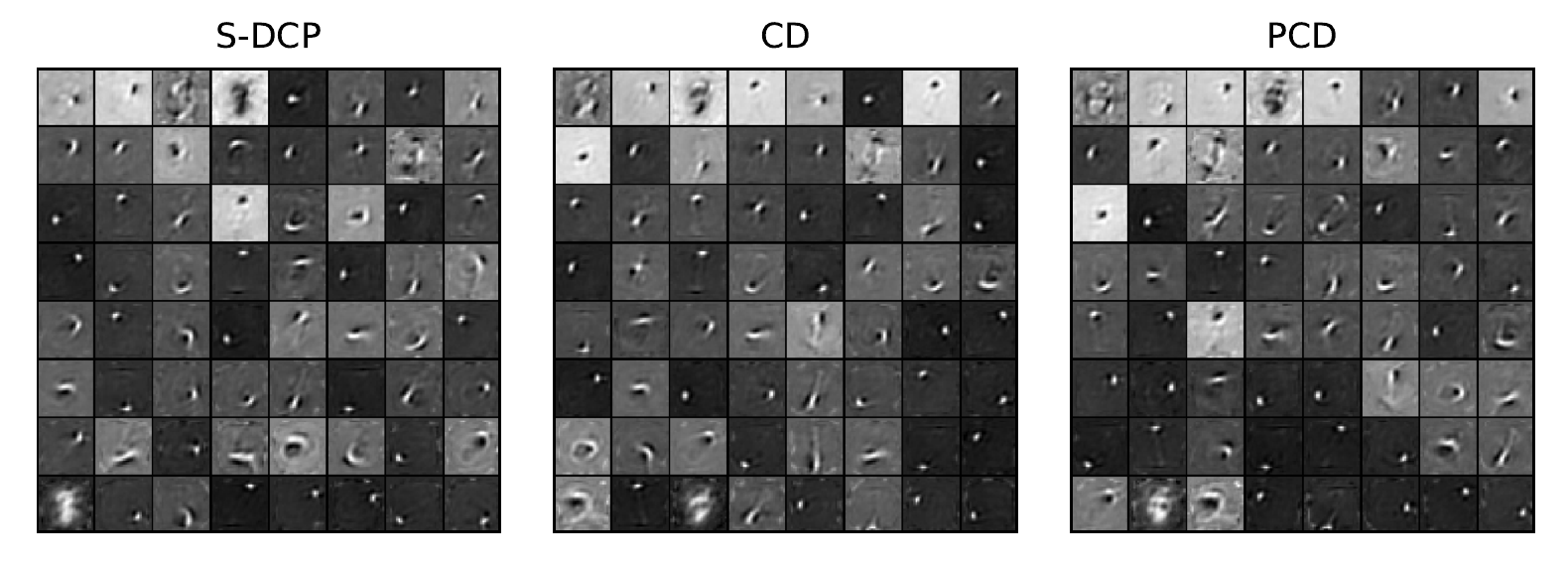}\label{fig:weights_mnist}
 \includegraphics[width=0.59\columnwidth]{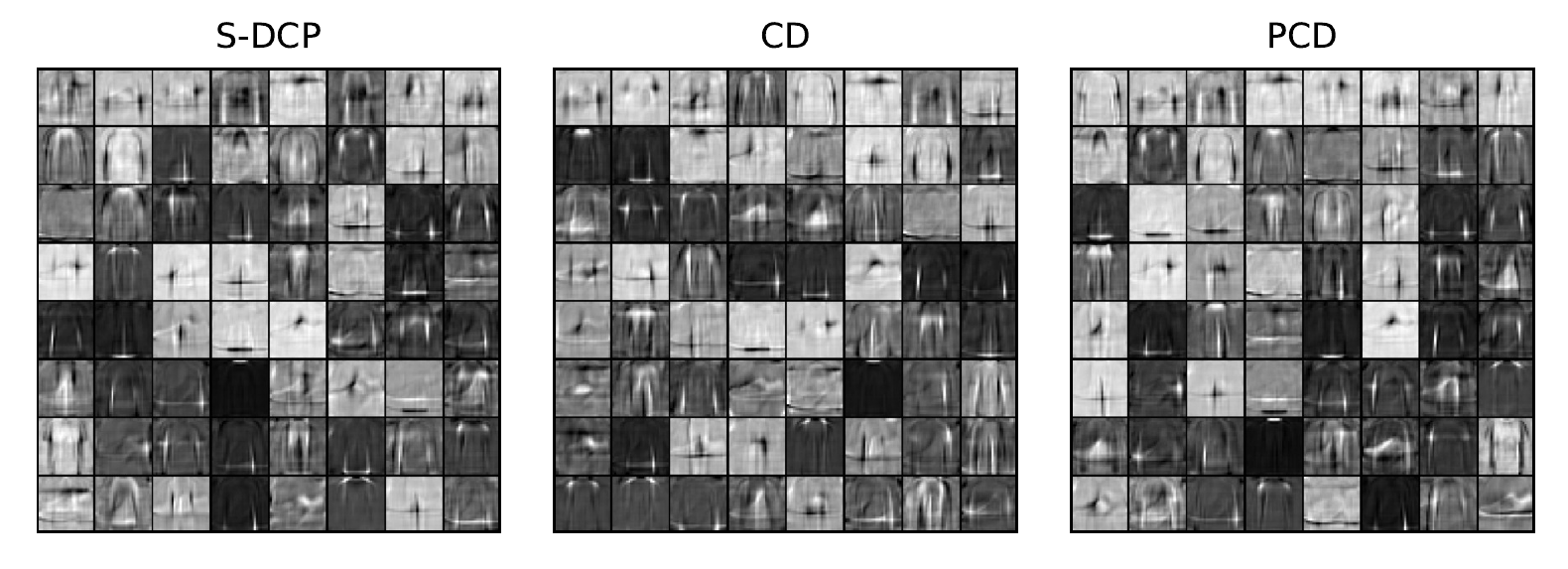}\label{fig:weights_fmnist}
  \includegraphics[width=0.59\columnwidth]{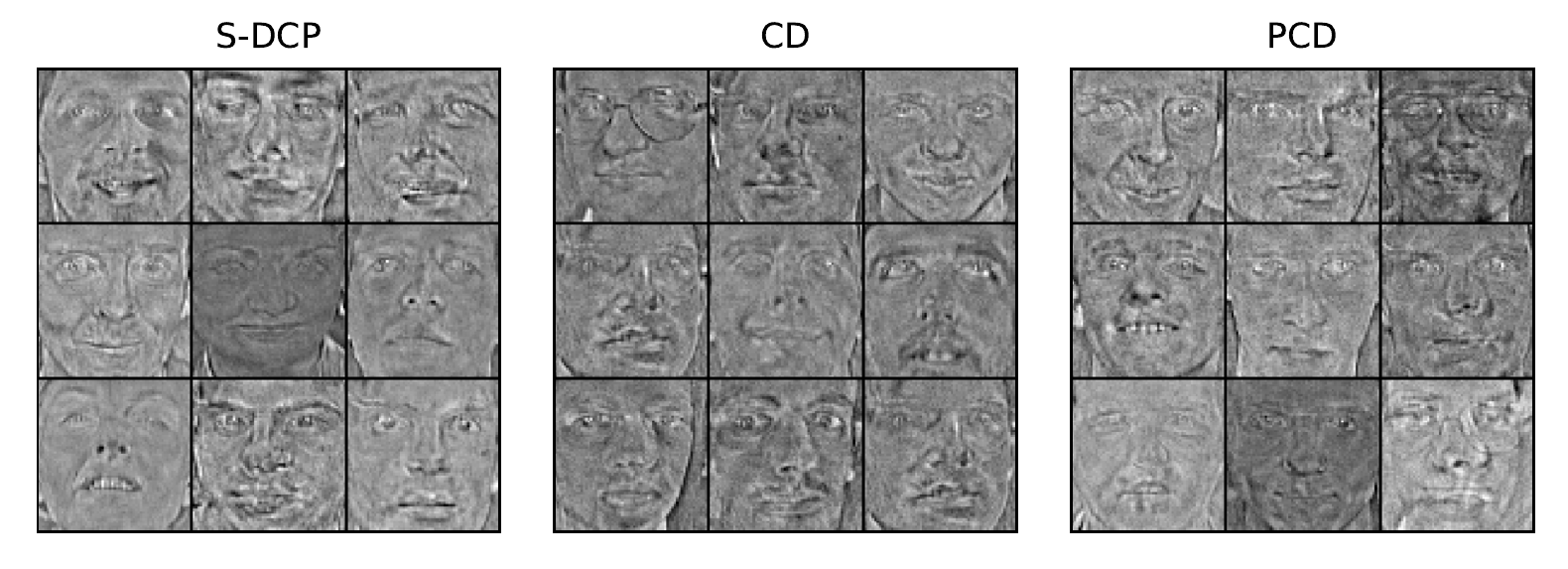}\label{fig:weights_olivetti}  
\caption{A few sample weights (obtained at the end of epochs at which the Ref. LL is achieved) or filters learnt using different algorithms on each dataset with fixed variance (For S-DCP, Algorithm \ref{Alg:S-DCP} is used) are plotted as grey-scale images.}\label{fig:filters_no_var}
\end{figure}

A few sample weights from the model learnt (at the end of epochs at which the Ref. LL is achieved) are plotted as filters in Fig. \ref{fig:filters_no_var}. To do so, the set of weights connected to each hidden unit (of dimension $m$) is
reshaped to the original image size and plotted as grey scale images. 

As observed from the Fig. \ref{fig:weights_natimages} and \ref{fig:weights_olivetti}, 
the weights are
able to capture the relevant  information from the training data.
Specifically, we observe that the filters capture the edge related information in the Natural Images dataset (\ref{fig:weights_natimages}) and extract the facial features from the  Olivetti faces dataset (\ref{fig:weights_olivetti}).
Thus, we find that S-DCP learns
filters comparable to those learnt by CD and PCD; but it learns them faster.

Models tend to learn point filters in the initial stages of training and 
slowly converge to edge-like filters when mean zero and unit variance normalization is used
instead of ZCA.
This phenomenon was observed in \cite{citeulike:7491128} on
natural images.
The same phenomenon is observed here in Fig. \ref{fig:weights_mnist} and Fig. \ref{fig:weights_fmnist}, for the MNIST and FMNIST datasets.
From the figures we find that most of the filters have only been able to capture the point-like information from
the data.
However, we observe that S-DCP results in more filters that capture edge-like
information compared to the CD and PCD algorithms. 
This indicates that the transition from point filters to
edge filters happens earlier with the S-DCP algorithm.

\subsubsection{Generative ability}\label{subsub:gen_ability}

\begin{figure}
\centering

  \includegraphics[width=0.59\columnwidth]{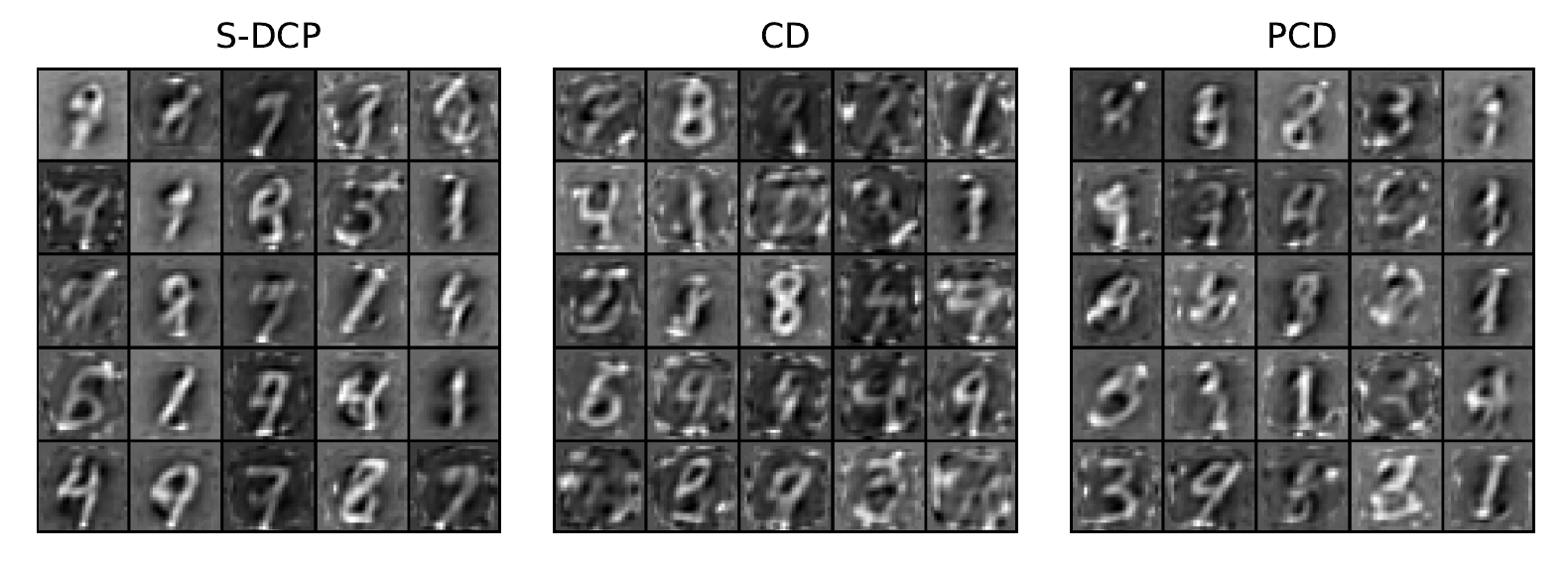}
 \includegraphics[width=0.59\columnwidth]{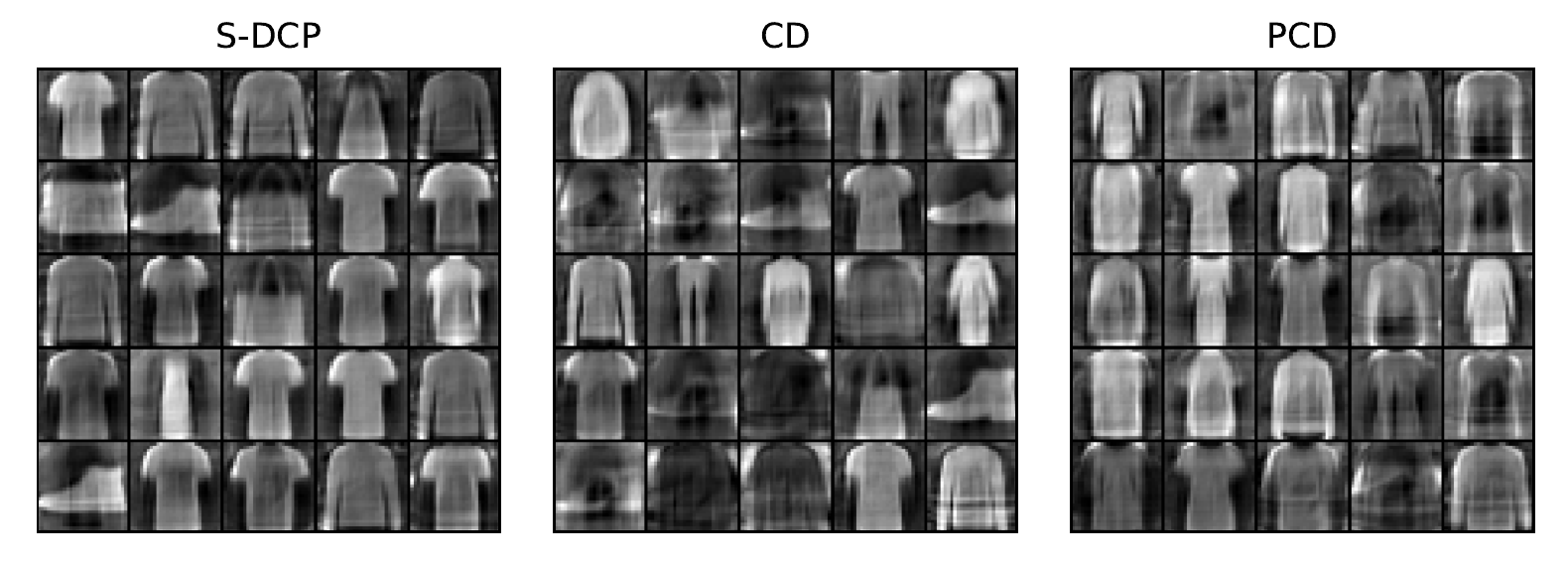}
\includegraphics[width=0.59\columnwidth]{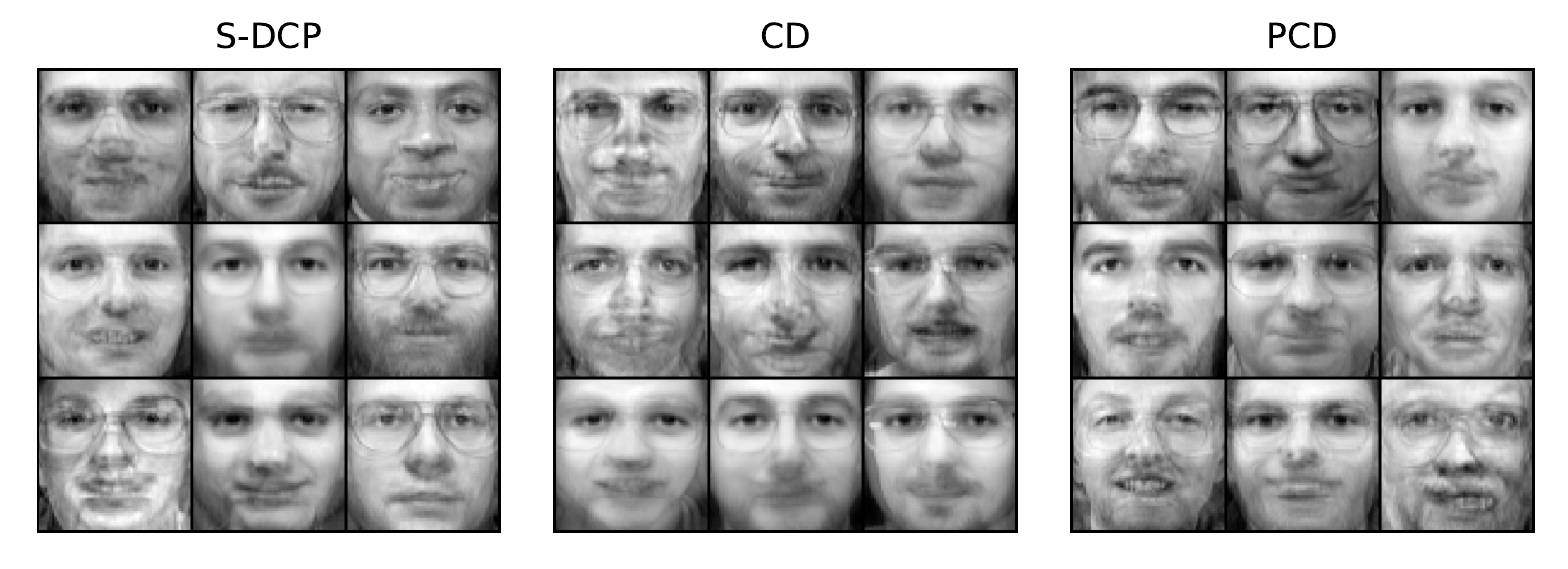}  
\caption{The samples from the models learnt (at the end of epochs at which the Ref. LL is achieved) using different algorithms on each dataset with fixed variance. For S-DCP, Algorithm
\ref{Alg:S-DCP} is used.}\label{fig:samples_no_var}
\end{figure}

The samples from the models learnt (at the end of epochs at which the Ref. LL is achieved) using different algorithms are given in Fig. \ref{fig:samples_no_var}. From the figure we observe that
the model learnt using S-DCP generates samples that are sharper and less noisy compared to those generated by the models learnt using the CD and PCD algorithms. 

\section{Conclusions}\label{sec:conclusions}
While GB-RBMs are good models for unsupervised representation learning, training them successfully is a challenging task.
The current standard algorithms namely, CD and PCD are very sensitive to the hyperparameters and hence, are not very efficient to train GB-RBMs.
In this work, we 
 showed that the negative log-likelihood of a GB-RBM is a difference of convex functions with respect to the weights, $w_{ij}$, and the biases of hidden units, $c_j$ (under the assumption that the variances of visible units are fixed) and this is utilized to employ the difference of convex functions programming approach to propose an efficient algorithm, called S-DCP.
We also proposed a variant of S-DCP where the learning of this variance is also facilitated.
Through extensive empirical results on multiple datasets, we showed that S-DCP results in
a considerable improvement in the speed of learning and the
quality of the final generative models learnt compared to those
of CD and PCD. The results presented in this study support the
claim that the S-DCP algorithm is currently a very efficient method to learn
GB-RBMs. In the S-DCP algorithm, the noisy gradient estimated through MCMC sampling is used for optimizing a convex function and this may be the reason why the optimization method is more effective compared to CD and PCD. Further work is needed in terms of theoretical analysis of such stochastic gradient descent on convex functions to better understand the optimization dynamics of S-DCP and to possibly further improve it. Another direction in which the work presented here can be extended is in terms of exploring the utility of second order methods based on estimated Hessian in the inner loop of the S-DCP which solves a convex optimization problem.

\appendix
\section{The function $f$ }\label{appendix_f_g}
\noindent Consider the partition function,
\beqa
Z(\theta)&=&\sum_\vech\int_\vecv e^{-\sum_i \frac{(v_i-b_i)^2}{2}+\sum_{i,j} w_{ij} v_i \, h_j +\sum_j c_j\,h_j} \mathbf{d}\vecv\non\\
&=&\sum_\vech e^{\sum_j c_j h_j+\sum_i \lb\sum_j w_{ij}\, h_j\rb\lb\frac{\sum_j w_{ij}\, h_j}{2}+b_i\rb} \times \non\\
&&\int_\vecv e^{-\sum_i \frac{\lb v_i-(\sum_j w_{ij} h_j+b_i)\rb^2}{2}} \mathbf{d}\vecv \non\\
&=& (2\pi)^{\frac{m}{2}}\sum_\vech e^{\sum_j c_j h_j+\sum_i \lb\sum_j w_{ij}\, h_j\rb\lb\frac{\sum_j w_{ij}\, h_j}{2}+b_i\rb}.\non
\eeqa
Therefore, $f(\theta)=\log\,\, Z(\theta)$ is given by,
\beq
f(\theta) = K_f+\, \log \sum_\vech e^{\sum\limits_j c_j h_j+\sum\limits_i \lb\sum\limits_j w_{ij}\, h_j\rb\lb\frac{\sum\limits_j w_{ij}\, h_j}{2}+b_i\rb}\non
\eeq
where, $K_f = \frac{m}{2}\log (2 \pi)$.
\section{Convexity of log-sum-exponential function}
\label{a_lse}
\noindent 
We first prove that $\it{lse}$ function ($\Real^d \rightarrow \Real$) given by,
\beq
lse(\vecu)=\log\sum\limits_i\beta_i \,\,e^{u_i}=\log \vecbeta^T e^{\vecu}
\label{lse_appendix}
\eeq
is convex \wrt $\vecu$. Here, $u_i$ denotes $i^\text{th}$ element of the vector $\vecu$ and $e^\vecu$ denotes element-wise exponential of $\vecu$.
The gradient and Hessian of $lse$ are given as, 
\beqa
\nabla \it{lse}(\vecu)&=&\frac{\vecbeta\odot e^{\vecu}}{\vecbeta^T e^{\vecu}}\non\\ 
\nabla^2 \it{lse}(\vecu)&=&\frac{\vecbeta^T e^{\vecu} \diag(\vecbeta\odot e^{\vecu})-(\vecbeta\odot e^{\vecu})(\vecbeta\odot e^{\vecu})^T}{(\vecbeta^T e^{\vecu})^2}\non\\
&=&\frac{ \diag(\vecbeta\odot e^{\vecu})}{\vecbeta^T e^{\vecu}} -  \lb\frac{\vecbeta\odot e^{\vecu}}{\vecbeta^T e^{\vecu}}\rb\lb\frac{\vecbeta\odot e^{\vecu}}{\vecbeta^T e^{\vecu}}\rb^T\non
\eeqa
where, $\odot$ denotes element-wise multiplication.
Let us denote, $\veca=\vecbeta\odot e^{\vecu}$. Now, the Hessian can be written as,
\beq
\nabla^2 \it{lse}(\vecu)=\frac{ \diag(\veca)}{\vecone^T \veca}-\frac{\veca \veca^T}{(\vecone^T \veca)^2}\non
\eeq
where, $\vecone$ is a vector of all 1's. Now we have,
\beqa
\vecx^T \nabla^2 \it{lse}(\vecu) \vecx &=& \vecx^T \lb \frac{\vecone^T \veca \,\diag(\veca)-\veca \veca^T}{(\vecone^T \veca)^2}\rb\vecx\non\\
&=& \frac{\sum a_k \sum a_k x_k^2-\lb \sum a_k x_k \rb^2}{(\sum a_k)^2}\ge 0.\non
\eeqa
Hence, the $\it{lse}$ function given in eq. \eqref{lse_appendix} is convex. Now, 
consider the function of the form (given in Lemma~\ref{lem_lse}),
\beq
\it{lse}_\veca(\vecu)=\log\sum\limits_{i=1}^N\beta_i \,\,e^{\veca_i^T \vecu}
\eeq
where, $\veca_1,\ldots,\veca_N$ are some fixed vectors in $\Real^d$.
Let $A$ be an $N \times d$ matrix whose rows are ${\bf a}_i$, then $\it{lse}_\veca({\vecs}) = \it{lse}(A{ \vecs}) .$
Now, for any two vectors $\vecs_1,\vecs_2\in \Real^d$ we have,
\beqa
\it{lse}_\veca(\alpha \vecs_1 + (1-\alpha)\vecs_2) & = & \it{lse}(A(\alpha \vecs_1 + (1-\alpha)\vecs_2)) \non\\
& \leq & \alpha \,\it{lse}(A\vecs_1) + (1-\alpha)\, \it{lse}(A\vecs_2) \non\\
& = & \alpha\, \it{lse}_\veca(\vecs_1) + (1 - \alpha)\, \it{lse}_\veca(\vecs_2).\non
\eeqa
Hence, the $\it{lse}_\veca$ function is convex which proves Lemma~\ref{lem_lse}.
\section{Convexity of log-sum-exponential-quadratic function}
\label{lemma_proof}
Consider the log-sum-exponential-quadratic function,
\beq
lse_q(\vecu)=\log\sum\limits_i \beta_i e^{\frac{1}{2}\vecu^T A_i \vecu+\veca_i^T \vecu}.\non
\eeq
The gradient of the above function is,
\beq
\nabla lse_q(\vecu)=\frac{\sum\limits_i \beta_i e^{\frac{1}{2}\vecu^T A_i \vecu+\veca_i^T \vecu} (A_i\vecu+\veca_i)}{\sum\limits_i \beta_i e^{\frac{1}{2}\vecu^T A_i \vecu+\veca_i^T \vecu}}.\non
\eeq
Let $\gamma(i)=\beta_i e^{\frac{1}{2}\vecu^T A_i \vecu+\veca_i^T \vecu}$ and $\bar{\gamma}(i)=\frac{\gamma(i)}{\sum_{i'} \gamma(i')}$. 
The Hessian of $\it{lse}_q(\vecu)$ is given as,
\beqa
\nabla^2 lse_q(\vecu) &{=}& \sum\limits_i \bar{\gamma}(i) A_i + \sum\limits_i \bar{\gamma}(i) (A_i\vecu+\veca_i) (A_i\vecu+\veca_i)^T \non\\
&&-\sum\limits_i \bar{\gamma}(i) (A_i\vecu+\veca_i) \sum\limits_i \bar{\gamma}(i) (A_i\vecu+\veca_i)^T.\non
\eeqa
Now, consider $\vecx^T \nabla^2 lse_q(\vecu) \vecx$,
\beq
\vecx^T \nabla^2 lse_q(\vecu) \vecx = \sum\limits_i \bar{\gamma}(i) \vecx^T A_i \vecx + \sum\limits_i \bar{\gamma}(i) \alpha_i^2 -\lb \sum\limits_i \bar{\gamma}(i) \alpha_i\rb^2\non
\eeq
where, $\alpha_i = \vecx^T (A_i\vecu+\veca_i)$.
By noting that $\sum\limits_i \bar{\gamma}(i) \alpha_i^2 -(\sum\limits_i \bar{\gamma}(i) \alpha_i)^2\ge 0$, we can write
\beq
\vecx^T \nabla^2 lse_q(\vecu) \vecx \ge \sum\limits_i \bar{\gamma}(i) \vecx^T A_i \vecx.\non
\eeq
If $A_i,\forall i$ is positive semi-definite then $\vecx^T \nabla^2 lse_q(\vecu) \vecx \ge 0$. Therefore, the log-sum-exponential-quadratic functions
of the considered form are convex.
\section*{Acknowledgment}

The authors thank the NVIDIA Corporation for the donation of the Titan X Pascal GPU used in this research.

\def\url#1{}
\def\doi#1{}

\end{document}